\documentclass[11pt]{article}

\makeatletter
    \renewcommand\section{\@startsection {section}{1}{\z@}%
                                       {-3.5ex \@plus -1ex \@minus -.2ex}%
                                       {2.3ex \@plus.2ex}%
                                       {\normalfont\fontfamily{phv}\fontsize{16}{19}\bfseries}}
    \renewcommand\subsection{\@startsection{subsection}{2}{\z@}%
                                         {-3.25ex\@plus -1ex \@minus -.2ex}%
                                         {1.5ex \@plus .2ex}%
                                         {\normalfont\fontfamily{phv}\fontsize{14}{17}\bfseries}}
    \renewcommand\subsubsection{\@startsection{subsubsection}{3}{\z@}%
                                        {-3.25ex\@plus -1ex \@minus -.2ex}%
                                         {1.5ex \@plus .2ex}%
                                         {\normalfont\normalsize\fontfamily{phv}\fontsize{14}{17}\selectfont}}
    \makeatother
\usepackage{amsmath}
\usepackage{amsthm} 
\usepackage{graphicx}
\usepackage{enumerate}
\usepackage{url}
\usepackage{textcomp}
\usepackage{xcolor}
\usepackage{float}
\usepackage{wrapfig}
\usepackage{multirow}
\usepackage{amsfonts}
\usepackage{mathrsfs}
\usepackage{amssymb,latexsym}
\usepackage{commath}
\usepackage{caption}
\usepackage{subcaption}
\usepackage{booktabs}
\usepackage{mathtools}
\usepackage{todonotes}
\usepackage{soul}
\usepackage{multicol}
\usepackage{amsbsy}
\usepackage{comment}
\usepackage[T1]{fontenc}
\usepackage[cmintegrals]{newtxmath}
\usepackage{bm}
\usepackage{array}
\usepackage{longtable}
\usepackage{setspace}
\usepackage{mdframed} 
\usepackage{hyperref}

\usepackage{algorithm}
\usepackage{algpseudocode}

\usepackage[round,authoryear]{natbib}
\let\cite\citep

\hypersetup{
	linkcolor=cyan,
	filecolor=magenta,
	urlcolor=blue,
}

\theoremstyle{definition} 

\newmdtheoremenv{theorem}{Theorem}
\newmdtheoremenv{lemma}{Lemma}
\newmdtheoremenv{proposition}{Proposition}
\newmdtheoremenv{definition}{Definition}
\newmdtheoremenv{property}{Property}
\newmdtheoremenv{assumption}{Assumption}

\graphicspath{{Figs/}}

\begin{document}

  \def\spacingset#1{\renewcommand{\baselinestretch}{#1}\small\normalsize} \spacingset{1}
          
  \title{P-K-GCN: Physics-augmented Koopman-enhanced Graph Convolutional Network for Deep Spatiotemporal Super-resolution}

  \author{Xizhuo(Cici) Zhang$^a$, Zekai Wang$^b$, Fei Liu$^c$, and Bing Yao$^a$\footnote{Corresponding author: Bing Yao, byao3@utk.edu}  \\  $^a$Department of Industrial \& Systems Engineering, \\The University of Tennessee, Knoxville, TN,  USA, 37996 \\
  $^b$ Charles F. Dolan School of Business, Fairfield University, Fairfield, USA\\
  $^c$Department of Electrical Engineering \& Computer Science, \\The University of Tennessee, Knoxville, TN,  USA, 37996 }

  \maketitle

\begin{abstract}
High-fidelity simulation of spatiotemporal dynamics is computationally prohibitive, necessitating efficient super-resolution techniques to reconstruct high-resolution data from coarse-grained inputs. Traditional data-driven methods often lack physical constraints, and simple physics-informed learning struggles with irregular spatial geometries and intricately evolving temporal dynamics.
To tackle these challenges, we propose a Physics-augmented Koopman-enhanced Graph Convolutional Network (P-K-GCN) for spatiotemporal super-resolution on irregular geometries. Specifically, a continuous spline-based GCN is first designed to extract spatial dependencies directly from coarse graph, and Koopman operator theory is incorporated to project the nonlinear dynamics into a compact latent space where temporal progression is linearized. Second, we augment the optimization objective with a physics-based loss to force the data-driven reconstructions to adhere to physical laws for improving predictive fidelity and robustness. Finally, we provide a rigorous theoretical analysis, establishing that the physics augmentation and Koopman regularization mathematically guarantees a reduction in super-resolution error by diminishing Rademacher complexity and tightening generalization bounds. We evaluate our framework on reconstructing spatially high-resolution cardiac electrodynamics across a 3D heart geometry from sparse low-resolution measurements. Numerical experiments demonstrate that our method achieves superior accuracy compared to baseline models. 

  \end{abstract}

  \noindent%

  \spacingset{1.5}


  \section{Introduction} \label{s:intro}

Spatiotemporal dynamics dictate essential processes across science and engineering, such as cardiac electrophysiology \citep{yang2023sensing} and environmental pollution tracking \citep{liu2023inverse}. These systems exhibit complex dependencies between spatial configurations and temporal evolution, typically characterized by nonlinear, high-dimensional dynamics. Modeling such systems at sufficiently fine spatiotemporal scales, i.e., high-resolution (HR), is essential for gaining detailed mechanistic insights and enabling reliable operational planning. For instance, high-fidelity simulations of cardiac electrodynamics provide unprecedented opportunities to evaluate treatment strategies and identify optimal ablation pathways for atrial fibrillation \citep{yao2025simulation}.

Despite the importance of HR representations, direct access to HR spatiotemporal information is rarely feasible in real-world applications. Practical measurement systems are often fundamentally constrained by limited sensor coverage, acquisition cost, and invasive accessibility. For example, multi-channel electrocardiogram recordings obtained from the cardiac surface are typically sparse, noisy, and restricted to limited spatial sampling locations \citep{chen2019characteristics, xie2022physics, yao2016physics}. Although high-fidelity numerical simulations can generate detailed HR solutions, accurately resolving complex nonlinear physical processes requires substantial computational resources and produces large-scale datasets, leading to prohibitive simulation, transmission, and storage costs \citep{ren2023physr}. As such, many practical workflows rely on spatiotemporally downsampled, low-resolution (LR) representations for storage and analysis. These limitations create a critical need for robust super-resolution (SR) methodologies capable of reconstructing physically consistent HR dynamics from LR observations.
However, achieving spatiotemporal SR of complex dynamic systems presents several fundamental challenges:

\textbf{(1) Ill-posed SR on complex geometries.} Spatiotemporal SR is an ill-posed inverse problem because multiple HR solutions may correspond to the same LR observation. Recovering fine-scale dynamics from sparse and noisy measurements is therefore intrinsically ambiguous, particularly in scientific systems governed by nonlinear physical processes \citep{chen2019characteristics, xie2022physics}. The challenge becomes more severe when the underlying dynamics evolve over complex irregular geometries. In many real-world applications, spatial domains are represented by unstructured meshes, curved manifolds, irregular surfaces, or other non-Euclidean structures, such as anatomical tissues and biological organs \citep{zhang2026geometry, yao2016mesh, calandra2016manifold, mak2018efficient, xie2024kronecker, chen2021function}. In such settings, the spatial relationships among measurement locations are highly heterogeneous and cannot be accurately characterized using regular-grid assumptions. Accurate SR requires geometry-aware representations capable of recovering missing HR details while respecting the underlying mesh topology and anatomical structure.

\textbf{(2) Temporally consistent SR under nonlinear dynamics.}
Unlike static image SR, spatiotemporal SR must reconstruct HR sequences that are not only accurate at each individual time step but also dynamically coherent across time. Complex physical systems often exhibit nonlinear and potentially unstable temporal evolution caused by environmental, biological, and physical interactions \citep{liu2018statistical, yao2021spatiotemporal}. In cardiac electrophysiology, for example, electrical impulse propagation across excitable media can generate reentrant spiral waves and spatiotemporal chaos associated with arrhythmogenesis \citep{wijesurendra2019mechanisms,xie2024automated}. If SR is performed independently at each frame, small reconstruction errors may accumulate over time, producing temporally inconsistent HR dynamics. Thus, effective SR models must recover fine-scale spatial details while preserving the latent temporal evolution of the underlying physical system.

\textbf{(3) Physical fidelity beyond statistical accuracy for SR.}
Many spatiotemporal processes are governed by well-established physical laws, such as reaction-diffusion PDEs for cardiac electrical propagation \citep{yang2023sensing, trayanova2011whole, zhang2025physics} and heat conduction equations for transient temperature evolution in additive manufacturing \citep{guo2022deep}. Purely data-driven SR methods may generate numerically accurate reconstructions that nevertheless violate physical constraints, especially under sparse-data or out-of-distribution settings \citep{karpatne2017theory}. This issue is particularly critical in safety-sensitive applications such as cardiac modeling, where physically inconsistent HR reconstructions may lead to unreliable interpretation. Therefore, robust SR requires physics-augmented learning mechanisms that constrain the LR-to-HR mapping toward physically admissible solutions.

To address these challenges, we propose a Physics-augmented Koopman-enhanced Graph Convolutional Network (P-K-GCN) tailored for the deep SR of spatiotemporal systems. The specific contributions are as follows:

\textbf{(1) Geometry-Aware Graph Convolution for Spatial Modeling:} The foundation of the proposed architecture operates on graph-based representations of LR irregular spatial domains. By utilizing continuous graph convolutions, the network effectively captures complex spatial dependencies and extracts local geometric features. The learned geometry-aware latent representations are subsequently decoded to reconstruct the corresponding HR spatial fields while preserving geometric continuity and anatomically meaningful structures.
    
\textbf{(2) Koopman-Enhanced Dynamics Learning for Temporal Consistency:} 
To model the nonlinear temporal evolution of spatiotemporal systems, the proposed framework incorporates Koopman operator theory into the latent representation space learned by the GCN encoder. Specifically, nonlinear system dynamics are projected into a compact latent space in which temporal evolution can be approximated by linear operators. This Koopman-based linearization enables stable temporal modeling, and the latent Koopman dynamics act as a temporal regularizer that enforces dynamical consistency across reconstructed HR frames, thereby improving temporal coherence, stability, and predictive robustness.
    
\textbf{(3) Physics-Augmented Regularization for Physical Fidelity:} 
To ensure that reconstructed HR solutions remain physically plausible, our framework further incorporates governing physical principles into the learning objective as soft constraints. This formulation regularizes the solution space toward physically admissible reconstructions that satisfy underlying physics constraints, thereby improving reconstruction fidelity and suppressing nonphysical artifacts under sparse sensing and limited-data conditions.

\textbf{(4) Theoretically-Grounded Bounds for Error Mitigation:} To provide a rigorous mathematical foundation, the designed architecture is supported by definitive theoretical guarantees for error mitigation. Physics augmentation restricts the hypothesis space to physically valid mappings, which reduces Rademacher complexity and strictly bounds the reconstruction error. The Koopman operator linearizes the latent space, capping the expansion rate to suppress the temporal error explosion typical of standard recurrent networks. These components all guarantee tighter SR bounds, ensuring the framework yields inherently superior SR performance.

We evaluate our framework on reconstructing spatially HR cardiac electrodynamics across a 3D heart geometry from sparse LR measurements. Experimental results and ablation studies demonstrate that P-K-GCN consistently outperforms state-of-the-art baselines in reconstruction accuracy and robustness, validating its effectiveness for physics-constrained SR on complex domains.



  \section{Research background}  \label{s:re-review}

\subsection{Spatiotemporal Super-resolution (SR)}

SR aims to reconstruct HR information from LR observations and has been widely studied as an ill-posed inverse problem. 
Early SR approaches primarily relied on sparse signal and dictionary-based representations, where HR image patches were reconstructed by exploiting sparse correspondences between LR and HR feature spaces \citep{yang2010image}. Although these methods demonstrated the feasibility of recovering fine-scale structures from coarse observations, their performance was often limited by handcrafted priors and restricted representational capacity.
With the rapid development of deep learning, convolutional neural networks (CNNs) substantially improved SR performance by learning nonlinear mappings from LR images to HR outputs directly from data \citep{dong2015image}. Subsequent studies further enhanced reconstruction quality through deeper residual learning, generative adversarial training, attention mechanisms, and sub-pixel convolution operations \citep{lim2017enhanced, shi2016real, zhang2018image}.

Extending SR from static images to dynamic visual data introduces the additional challenge of preserving temporal coherence while recovering fine-scale spatial details. Video SR must recover both spatial details and temporal consistency. Existing methods commonly combine spatial upsampling with motion estimation, frame interpolation, or recurrent convolutional architectures to capture temporal dependencies across consecutive frames \citep{niklaus2017video, jiang2018super, sajjadi2018frame}. Representative spatiotemporal deep architectures, such as ConvLSTM, extend recurrent neural networks with convolutional operators, where convolutional layers extract local spatial patterns within each frame and recurrent units model temporal evolution across frames \citep{shi2015convolutional}. Such CNN- and ConvLSTM-based models have been widely adopted for structured spatiotemporal prediction tasks, including urban planning, transportation systems, and computer vision \citep{majidizadeh2024semantic, wang2020spatio}.


In scientific and engineering domains, SR methods have been increasingly employed to recover fine-scale dynamics from LR observations generated by sparse measurements, coarse computational discretizations, or incomplete experimental sampling.
Recent studies have demonstrated the effectiveness of deep learning-based SR frameworks in recovering unresolved structures in turbulence modeling, fluid dynamics, smoke simulation, and climate systems \citep{xie2018tempogan, liu2020deep, fukami2021machine, stengel2020adversarial}. However, unlike natural image SR, scientific SR imposes substantially more stringent reconstruction requirements, as the recovered HR fields must not only achieve agreement with reference observations but also remain consistent with governing physical laws, boundary conditions, and conservation principles of the spatiotemporal systems.

To incorporate such prior knowledge, recent studies have explored physics-informed SR for dynamical systems governed by partial differential equations (PDEs). These methods augment data-fitting objectives with physics-based losses, hard or soft boundary-condition constraints, or differentiable numerical operators so that the reconstructed HR solutions remain consistent with the underlying physical model \citep{wang2020physics, gao2021super}. Existing physics-informed SR approaches can be broadly categorized into spatial SR, which reconstructs fine spatial fields from coarse snapshots \citep{wang2020physics, gao2021super, subramaniam2020turbulence}; temporal SR, which synthesizes missing intermediate states \citep{ren2022phycrnet, ren2023physr}; and spatiotemporal SR, which simultaneously enhances spatial and temporal resolution \citep{esmaeilzadeh2020meshfreeflownet, ren2023physr}.  Nevertheless, most existing CNN-based and physics-informed SR methods are built around convolutional or recurrent operators on uniform Cartesian grids. This assumption limits their direct applicability to irregular domains, such as curved biological surfaces and unstructured finite-element meshes, where spatial connectivity is non-Euclidean and heterogeneous. 

\subsection{Geometry-aware Spatiotemporal Neural Networks}


GCNs and graph neural networks (GNNs) generalize convolution to graph-structured data using adjacency matrices, graph Laplacians, or message-passing operators \citep{bronstein2017geometric, wang2025transformer}. This formulation enables neural networks to aggregate information from neighboring nodes while respecting irregular spatial connectivity. As a result, graph-based models provide a natural representation for non-Euclidean topologies. To model evolving dynamical systems on irregular geometries, spatiotemporal GNN (STGNNs) integrate graph-based spatial representation learning with temporal evolution modules such as gated temporal convolutions, recurrent units, or attention mechanisms \citep{yu2017spatio, ali2022exploiting, wen2023diffstg}. These models have shown strong potential for forecasting and spatiotemporal representation learning on graph-structured data. Nevertheless, many existing approaches remain limited in modeling long-horizon nonlinear dynamics due to temporal instability, error accumulation, and the difficulty of learning highly nonlinear evolution operators directly in the observation space.


 Koopman operator theory provides an alternative dynamical systems perspective by transforming nonlinear system evolution into approximately linear dynamics within a learned latent observable space \citep{lusch2018deep, takeishi2017learning, wang2025transformer}. In this framework, nonlinear temporal behavior is represented through linear evolution operators acting on latent observables, enabling more stable long-term propagation and improved interpretability of dynamical evolution. When integrated with graph-based spatial encoders, Koopman-enhanced latent dynamics learning offers a promising mechanism for simultaneously capturing non-Euclidean spatial dependencies and temporally coherent system evolution in spatiotemporal scientific applications.


A growing body of work has further incorporated physical knowledge into graph learning on meshes and irregular domains. Graph-based physical simulators use graph connectivity to represent interactions among particles or mesh elements. Sanchez-Gonzalez et al. proposed Graph Network-based Simulators (GNS), where particles are treated as graph nodes and physical interactions are learned through message passing \citep{sanchez2020learning}. This framework successfully models complex systems such as fluids, rigid bodies, and deformable materials, demonstrating the ability of graph networks to capture local physical interactions on irregular particle systems. Pfaff et al. further extended this idea to mesh-based physical simulation through MeshGraphNets, which perform message passing over mesh connectivity and support adaptive mesh representations \citep{pfaff2021learning}. However, both GNS and MeshGraphNets are primarily designed for forward time evolution from known system states. They do not explicitly address the inverse problem of recovering HR spatiotemporal fields from sparse or coarse graph LR observations.

Physics-informed GNNs incorporate physical laws or numerical operators into graph learning. Zhang et al. proposed a Physics-Informed GNN (PIGNN), which combines GNN-based representation learning with physical constraints and finite-difference approximations for forward and inverse nonlinear PDE problems \citep{zhang2024pignn}. By replacing coordinate-based neural approximation with graph-based discrete operators, PIGNN extends PINN framework to irregular structures. Zeng et al. introduced PhyMPGN, a physics-encoded message-passing graph network for spatiotemporal PDE systems on irregular meshes \citep{zeng2025phympgn}. PhyMPGN embeds graph message passing within a numerical integration framework, introduces a learnable Laplace block to encode diffusion-related physics, and uses boundary-condition padding to improve prediction accuracy. These methods show that physical constraints can improve graph-based PDE learning on irregular domains. Nevertheless, their main objective remains PDE solution approximation, parameter inference, or forward temporal forecasting. They do not directly solve the ill-posed LR-to-HR reconstruction problem required by spatiotemporal SR.

Geometry-aware PINNs and physics-constrained learning methods extend mechanistic regularization to complex domains. Costabal et al. proposed $\Delta$-PINNs, which use Laplace--Beltrami eigenfunctions as geometry-aware positional encodings and finite-element operators to represent PDE constraints on complex geometries \citep{costabal2024delta}. This method enables PINNs to account for intrinsic geometric distances and domain topology, improving performance when standard Euclidean coordinates are insufficient. In cardiac modeling, physics-constrained deep learning and active learning frameworks have incorporated electrophysiological PDE priors to infer cardiac electrodynamics from sparse observations \citep{yao2024multi, xie2022physics2}. However, they are mainly formulated as coordinate-based PDE solvers, and do not provide explicit graph-to-graph resolution enhancement.

  \section{Research Methodology}
\label{s:method}

\begin{figure}[!ht]
  \centering
  \includegraphics[width=0.9\linewidth]{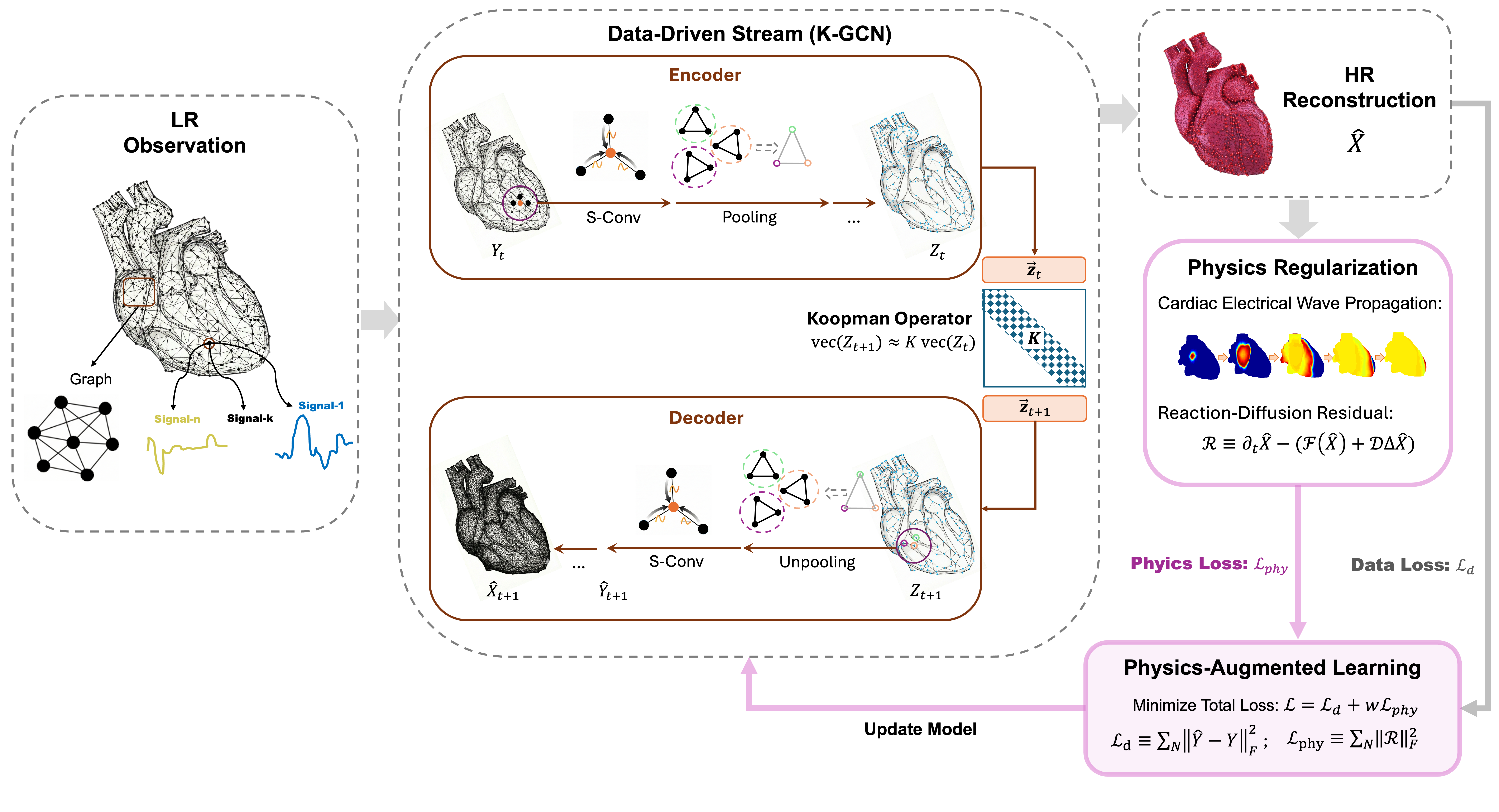}
  \caption{Flowchart of the proposed methodology for spatiotemporal super-resolution. 
  }
  \label{fig:pkgnn_flowchart}
\end{figure}

  Fig.~\ref{fig:pkgnn_flowchart} illustrates the overall architecture of our P-K-GCN framework. Let $\boldsymbol{q}(\boldsymbol{x},t) \in \mathbb{R}^{C}$ denote the system state at spatial location $\boldsymbol{x}$ and time $t$, where $C$ is the number of feature channels. The available LR observations are acquired at sparsely distributed spatial locations $\mathcal{X}_l=\{\boldsymbol{x}_1,\dots,\boldsymbol{x}_{N_\text{s}}\}$ and over a discrete set of time instances $\mathcal{T}=\{t_1,\dots,t_{N_\text{t}}\}$, thereby forming the tensor $\tilde{\mathcal{Q}}_l \in \mathbb{R}^{C \times N_\text{s} \times N_\text{t}}$. This observed tensor is the sum of the underlying noise-free data tensor $\mathcal{Q}_l$ and a measurement noise term $\mathbf{\Xi}_\text{mea}$.
  The objective of spatiotemporal SR is to reconstruct the corresponding HR dynamics $\mathcal{Q}_h \in \mathbb{R}^{C \times N_\text{s}^* \times N_\text{t}}$, across a dense set of spatial locations $\mathcal{X}_h=\{\boldsymbol{x}_1,\dots,\boldsymbol{x}_{N_\text{s}^*}\}$. To achieve this, the proposed framework first employs a graph-based encoder to map the LR observations $\tilde{\mathcal{Q}}_l$ defined on the irregular geometry into a compact latent representation space that preserves the underlying geometric structure. The latent dynamics are subsequently propagated through a Koopman operator formulation, where the nonlinear temporal evolution is approximated by linear dynamics in the latent observable space, enabling stable modeling across the observation horizon. A decoder then performs spatial refinement and resolution upsampling to reconstruct the HR spatiotemporal fields $\mathcal{Q}_h$. To further improve reconstruction robustness, the training objective incorporates physics-based regularization derived from the governing physical laws, thereby constraining the reconstructions to remain physically consistent with the underlying dynamical system. (A detailed notation table is available in Supplementary Material~\ref{app:notations}.)



\subsection{Geometry-Aware Graph Convolution Modeling}

To effectively extract spatial features, we model the 3D complex geometry as an undirected graph, denoted by $\mathcal{G} = (\mathcal{V}, \mathcal{E}, \mathbf{W}, \boldsymbol{q})_{\mathcal{E} \subseteq \mathcal{V} \times \mathcal{V}, \mathbf{W} \in [0,1]^{N \times N \times 3}}$. Here, $\mathcal{V}$ represents the set of $N$ vertices, while the adjacency matrix $\mathcal{E}$ defines the binary connectivity between any two nodes $i$ and $j$. The geometric relationships are captured by the edge attribute tensor $\mathbf{W}$, where each non-zero element $\boldsymbol{w}(i,j) \in \mathbb{R}^3$ indicates the normalized spatial displacement between connected vertices $i$ and $j$ with $\mathcal{E}_{ij} = 1$. The SR mapping of a spatial snapshot from LR to HR graph domain at a specific time step $t$ is then formulated within a graph-based Encoder-Decoder framework:

\begin{equation}
    \mathcal{G} \left(\widehat{\mathbf{Q}}_h(t)\right) = \mathrm{D}_\theta \Bigl( \mathrm{E}_\theta \bigl( \mathcal{G} \left(\tilde{\mathbf{Q}}_l(t) :=\mathbf{Q}_l(t)+\mathbf{\Xi}_\text{mea}(t)\right)\bigr) \Bigr)
\end{equation}
where $\widehat{\mathbf{Q}}_h(t)$ is the reconstructed HR field and $\tilde{\mathbf{Q}}_l(t)$ is the LR observation.
This graph learning framework is defined based on continuous Spline Convolutiona (S-Conv) with residual connections and hierarchical graph coarsening and refinement, which are detailed as follows.

\paragraph{Continuous Spline Convolution (S-Conv).}
Following the SplineCNN approach \citep{fey2018splinecnn, wang2025transformer}, we leverage the local geometric information encoded in the edge attributes $\mathbf{W}$ to parameterize the adaptive convolutional kernels. Specifically, the edge-wise pseudo-coordinates $\boldsymbol{w}(i,j)$ act as dynamic controllers of how neighboring node features are aggregated, turning message passing into a geometry-aware process that fully exploits the structure of the graph. 
To achieve the geometric adaptivity, we employ continuous B-spline kernel functions, which parameterize convolutional weights as smooth, differentiable functions of the pseudo-coordinates. In particular, the kernel is constructed from tensor-product basis functions defined over a $D$-dimensional pseudo-coordinate space (with $D=3$ for 3D geometries). For each coordinate dimension $i \in \{1,\dots,D\}$, we define a uniform knot vector and specify the kernel resolution via the vector $\boldsymbol{r} = [r_1, \dots, r_D]^\top$, where $r_i$ defines the number of basis functions along the $i$-th dimension. Let $\mathcal{P} = \{1, \dots, r_1\} \times \cdots \times \{1, \dots, r_D\}$ denote the Cartesian product of these basis indices across all $D$ dimensions, forming the basis grid. For a given edge attribute vector $\boldsymbol{w} = [w_1, \dots, w_D]^\top \in \mathbf{W}$, 
and a basis index $\boldsymbol{p} = [p_1, \dots, p_D]^\top \in \mathcal{P}$, the B-spline basis function is formulated as:
\begin{equation}
    B_{\boldsymbol{p}}(\boldsymbol{w}) = \prod_{i=1}^D N_{i,p_i}(w_i)
\end{equation}
where $N_{i,p_i}(w_i)$ represents the 1D B-spline basis function parameterized by the index $p_i$ and evaluated at the scalar coordinate $w_i$.

To further translate the geometric relationships into convolution weights, the graph convolutional kernel $\mathbf{G}_{\mathbf{\Theta}}(\boldsymbol{w})$ is constructed as a continuous function of the edge attributes. The convolution maps input node features of dimension $C_\text{in}$ to output features of dimension $C_\text{out}$, which is formulated as a linear combination of the scalar basis products weighted by a set of trainable parameter matrices $\mathbf{\Theta} = \{\mathbf{\Theta}_{\boldsymbol{p}}\}_{\boldsymbol{p} \in \mathcal{P}}$, $\mathbf{\Theta}_{\boldsymbol{p}} \in \mathbb{R}^{C_\text{out} \times C_\text{in}}$:
\begin{equation}
    \mathbf{G}_{\mathbf{\Theta}}(\boldsymbol{w}) = \sum_{\boldsymbol{p} \in \mathcal{P}} \mathbf{\Theta}_{\boldsymbol{p}} B_{\boldsymbol{p}}(\boldsymbol{w})
\end{equation}

Finally, the continuous graph convolution between the input node features $\boldsymbol{q}$ (vectorized $\mathbf{Q}+\mathbf{\Xi}_\text{mea}$) and the learned kernel $\mathbf{G}_{\mathbf{\Theta}}(\cdot)$ is formulated as an aggregation over the local geometric neighborhood. For node $i$, the updated feature vector is calculated by applying the evaluated kernel matrix to the features of its local geometric neighborhood $\mathcal{N}(i)$:
\begin{equation}
    (\boldsymbol{q} * \mathbf{G})_i = \frac{1}{|\mathcal{N}(i)|} \sum_{j \in \mathcal{N}(i)} \mathbf{G}_{\mathbf{\Theta}}(\boldsymbol{w}(i,j)) \boldsymbol{q}_j
\end{equation}
where $\boldsymbol{q}_j \in \mathbb{R}^{C_\text{in}}$ is the input feature vector of node $j$. 


\paragraph{Residual Spatial Feature Extraction. } We further incorporate a residual connection to stabilize training and improve feature propagation within our spatial block (visualized in Fig. \ref{Fig:NN_details}). Let $\boldsymbol{q}^{(k)}$ denote the input features at layer $k$, which are processed by an S-Conv layer to extract localized geometric information. Simultaneously, $\boldsymbol{q}^{(k)}$ is propagated through a parallel residual pathway utilizing a $1\times1$ convolution (F-Conv) to map the features into the desired dimensional space. The outputs of both pathways are passed through non-linear activation functions, combined via element-wise addition, and subjected to a final activation. This composite structure facilitates effective gradient flow and mitigates gradient degradation \citep{he2016deep}. With $\rho(\cdot)$ denoting the ELU activation, the residual operation is mathematically defined as:
\begin{equation}
    \boldsymbol{q}^{(k+1)} = \rho \left( \rho \left( \operatorname{S-Conv} \left( \boldsymbol{q}^{(k)} \right) \right) + \rho \left( \operatorname{F-Conv} \left( \boldsymbol{q}^{(k)} \right) \right) \right)
    \label{eq:resblock}
\end{equation}

\paragraph{Hierarchical Graph Coarsening and Refinement.} We employ a hierarchical graph coarsening strategy to capture multi-scale spatial features. Specifically, we first group the nodes using the Graclus algorithm \citep{dhillon2007weighted} to extract intrinsic spatial features of the LR graph $\mathcal{G}_l$, yielding a hierarchy of structurally simplified meshes. The clustering topology at any given coarsening level $k$ is then encoded within a binary matrix $\mathbf{P}_k$. Functionally, an element $(\mathbf{P}_k)_{ij} = 1$ dictates that the $j$-th node from the finer graph is a member of the $i$-th coarsened cluster; otherwise, it is zero.
The node features for the subsequent, lower-resolution level $k+1$ are computed by averaging the traits of the grouped nodes:
\begin{equation}
    \mathbf{Q}_{k+1}^\text{pool} = \boldsymbol{\Delta}_{k+1} \mathbf{P}_k \mathbf{Q}_k
\end{equation}
where diagonal matrix $\boldsymbol{\Delta}_{k+1}$ acts as a spatial normalization factor with diagonal elements defined as $(\boldsymbol{\Delta}_{k+1})_{ii} = 1 / \sum_{j} (\mathbf{P}_k)_{ij}$. This operation yields multi-resolution feature representations that encode progressively larger receptive fields while maintaining consistency with the underlying graph topology.

For reconstruction of HR graph signals, the decoder $\mathrm{D}_\theta$ performs a symmetric hierarchical refinement process that reverses the coarsening procedure. At each stage, the coarse feature maps are unpooled by multiplying them with the transpose of the binary cluster assignment matrix, $\mathbf{P}_k^\text{T}$:
\begin{equation}
    \mathbf{Q}_k^\text{unpool} = \mathbf{P}_k^\text{T} \mathbf{Q}_{k+1}^\text{pool}
\end{equation}
This operation propagates each coarse node embedding to all fine nodes within its associated cluster, thereby preserving the hierarchical assignment structure induced during encoding.





\subsection{Koopman-Enhanced Learning of Nonlinear Dynamics}
  The graph-based Encoder–Decoder architecture induces a compact low-dimensional latent representation that provides an efficient space for temporal dynamics modeling by mitigating the curse of dimensionality inherent in the spatiotemporal domain. Building upon this representation, we adopt an operator-theoretic framework based on Koopman theory, wherein the nonlinear spatiotemporal dynamics are approximated by a linear evolution operator acting on latent observables. Specifically, after the final graph coarsening and pooling stage, the resulting latent graph $\mathcal{G}_z$ comprises $N_{s,z}$ nodes with $C_z$-dimensional feature embeddings. The corresponding latent state at time $t$ is defined as $\mathbf{Z}(t) = \mathrm{E}_\theta(\tilde{\mathbf{Q}}_l(t)) \in \mathbb{R}^{N_{s,z} \times C_z}$.  
  

  The latent state is then flattened into an observable state vector:
$
    \boldsymbol{z}(t) = \operatorname{vec}\bigl(\mathbf{Z}(t)\bigr) \in \mathbb{R}^{d}
    \quad \text{wher,e} \quad d = N_{s,z} C_z
$.
Building upon this vectorized representation, we introduce a trainable matrix $\mathbf{K} \in \mathbb{R}^{d \times d}$ as a finite-dimensional approximation of the Koopman operator. 
Although the Koopman operator is theoretically defined on an infinite-dimensional space of observables~\citep{koopman1931hamiltonian, brunton2022modern}, practical implementations typically rely on finite-dimensional approximations to enable tractable computation. Existing approaches construct such approximations either through predefined dictionary-based projections~\citep{williams2015data}, or through jointly learned neural encoders coupled with linear latent evolution operators~\citep{lusch2018deep, takeishi2017learning}. These formulations provide computationally efficient surrogates for modeling high-dimensional nonlinear dynamical systems.
By treating $\mathbf{K}$ as a trainable parameter, we can learn a linear latent transition model that captures the essential nonlinear evolution of the physical system. Specifically, this matrix $\mathbf{K}$ linearly advances the latent state in time:
\begin{equation}
    \boldsymbol{z}^K(t+1) \approx \mathbf{K} \boldsymbol{z}(t)
\end{equation}

To recover the spatial structure, $\boldsymbol{z}^K(t+1)$ is reshaped back into matrix form via $\widehat{\boldsymbol{Z}}^K(t+1) = \operatorname{vec}^{-1}\!\bigl(\boldsymbol{z}^K(t+1)\bigr)$. The decoder $\mathrm{D}_\theta$ subsequently maps this latent representation to the HR domain, yielding $\widehat{\mathbf{Q}}_h^K(t+1) = \mathrm{D}_\theta(\widehat{\mathbf{Z}}^K(t+1))$. 

\subsection{Physics-Augmented Model Optimization}

To further embed prior physical knowledge into the modeling framework and enhance SR fidelity, we formulate parameter inference as a physics-augmented optimization problem with the loss function defined:
\begin{equation}
    \mathcal{L} = \mathcal{L}_{\mathrm{d}} + w_{\mathrm{phy}} \, \mathcal{L}_{\mathrm{phy}}
    \label{eq:total_loss}
\end{equation}
where $\mathcal{L}_{\mathrm{d}}$ denotes the data-driven loss quantifying empirical data fit, and $\mathcal{L}_{\mathrm{phy}}$ enforces physical constraints derived from the governing equations, acting as a soft penalty modulated by the weight $w_{\mathrm{phy}}$. The network components, i.e., the encoder $\mathrm{E}_\theta$, decoder $\mathrm{D}_\theta$, and Koopman matrix $\mathbf{K}$, are optimized jointly with respect to $\mathcal{L}$. The comprehensive training procedure is outlined in Algorithm~\ref{alg:pk-gcn}.

\begin{algorithm}[ht]
\caption{P-K-GCN: Training Procedure}
\label{alg:pk-gcn}
\begin{algorithmic}[1]
\Require 
    LR observation block $\mathcal{B}_l = \{\tilde{\mathbf{Q}}_l(t+n)\}_{n=0}^{B-1}$,
    downsampling projection matrix $\mathbf{P}_{h \to l}$,
    temporal block size $B$,
    physics penalty weight $w_{\mathrm{phy}}$,
    learning rate $\eta$.
\Ensure 
    Optimized spatial encoder $\mathrm{E}_\theta$, decoder $\mathrm{D}_\theta$, and Koopman matrix $\mathbf{K}$

\State Initialize network parameters $\theta$ and Koopman matrix $\mathbf{K}$
\While{stopping criteria not met}
    
    \State \textbf{Phase 1: Spatial Encoding and Direct Reconstruction}
    \For{$n = 0, \dots, B-1$} 
        \State $\mathbf{Z}(t+n) \gets \mathrm{E}_\theta(\tilde{\mathbf{Q}}_l(t+n))$ \Comment{Encode to latent graph matrix}
        \State $\widehat{\mathbf{Q}}_h(t+n) \gets \mathrm{D}_\theta(\mathbf{Z}(t+n))$ \Comment{Reconstruct instantaneous HR spatial matrix}
    \EndFor

    \State \textbf{Phase 2: Koopman Latent Advancement and Prediction}
    \For{$n = 1, \dots, B-1$}
        \State $\boldsymbol{z}(t+n-1) \gets \operatorname{vec}(\mathbf{Z}(t+n-1))$ \Comment{Flatten to observable state vector}
        \State $\boldsymbol{z}^K(t+n) \gets \mathbf{K} \boldsymbol{z}(t+n-1)$ \Comment{Advance latent vector linearly}
        \State $\widehat{\mathbf{Z}}^K(t+n) \gets \operatorname{vec}^{-1}(\boldsymbol{z}^K(t+n))$ \Comment{Reshape back to latent matrix}
        \State $\widehat{\mathbf{Q}}_h^K(t+n) \gets \mathrm{D}_\theta(\widehat{\mathbf{Z}}^K(t+n))$ \Comment{Predict future HR spatial matrix}
    \EndFor

    \State \textbf{Phase 3: Loss Computation}
    \State Compute data-driven loss $\mathcal{L}_{\mathrm{d}}$ (Eq.~\ref{eq:data_loss_refined}) using $\mathcal{B}_l$, $\widehat{\mathbf{Q}}_h$, and $\widehat{\mathbf{Q}}_h^K$
    \State Compute physics loss $\mathcal{L}_{\mathrm{phy}}$ (Eq.~\ref{eq:phys_loss})
    \State Aggregate objective: $\mathcal{L} \gets \mathcal{L}_{\mathrm{d}} + w_{\mathrm{phy}} \, \mathcal{L}_{\mathrm{phy}}$

    \State \textbf{Phase 4: Backpropagation and Parameter Update}
    \State $\theta \gets \theta - \eta \nabla_\theta \mathcal{L}$ \Comment{Update Encoder $\mathrm{E}_\theta$ and Decoder $\mathrm{D}_\theta$}
    \State $\mathbf{K} \gets \mathbf{K} - \eta \nabla_{\mathbf{K}} \mathcal{L}$ \Comment{Update Koopman Operator $\mathbf{K}$}
\EndWhile
\State \Return $\mathrm{E}_\theta, \mathrm{D}_\theta, \mathbf{K}$
\end{algorithmic}
\end{algorithm}

\paragraph{Data-Driven Loss for LR Consistency.}
Given that training relies exclusively on sparse LR observations, we define a fixed projection matrix $\mathbf{P}_{h \to l} \in \mathbb{R}^{N_s^* \times N_s}$ to map the HR predictions back to the observational domain. The data-driven loss over a temporal block $\mathcal{B}_l$ of size $B$ is defined as:
\begin{equation}
    \mathcal{L}_{\mathrm{d}} = \sum_{n=0}^{B-1} \bigl\| \mathbf{P}_{h \to l}^{\top} \widehat{\mathbf{Q}}_h(t+n) - \tilde{\mathbf{Q}}_l(t+n) \bigr\|_F^2 + \sum_{n=1}^{B-1} \bigl\| \mathbf{P}_{h \to l}^{\top} \widehat{\mathbf{Q}}_h^K(t+n) - \tilde{\mathbf{Q}}_l(t+n) \bigr\|_F^2
    \label{eq:data_loss_refined}
\end{equation}
where the first term functions as a static reconstruction loss, penalizing deviations in the instantaneous spatial mapping $\widehat{\mathbf{Q}}_h(t+n)$ produced by the autoencoder; the second term is the Koopman dynamics loss, which evaluates the temporal predictions $\widehat{\mathbf{Q}}_h^K(t+n)$ derived from advancing the latent state via the Koopman operator before decoding. 


\paragraph{Physics Loss on Reconstructed HR Fields.}
This study focuses the SR of a reaction-diffusion system evolving on a 3D surface manifold $M$. The evolution dynamics is governed by the following PDEs:
\begin{equation}
    \begin{aligned}
        \frac{\partial u}{\partial t} &= e_1 \Delta u + g_1(u,v), \\
        \frac{\partial v}{\partial t} &= e_2 \Delta v + g_2(u,v),\\
        \mathbf{n}\cdot \nabla u|_M &= 0, \\
        \mathbf{n}\cdot \nabla v|_M &= 0
    \end{aligned}
    \label{eq:pdes}
\end{equation}
where $u$ and $v$ denote the spatiotemporal state variables, $\Delta$ represents the spatial Laplacian operator, $e_1$ and $e_2$ are the respective diffusion coefficients, $g_1(\cdot, \cdot)$ and $g_2(\cdot, \cdot)$ characterize the nonlinear reaction kinetics, the Neumann boundary conditions, expressed in terms of the outward unit normal vector $\mathbf{n}$, enforce a zero-flux constraint across the domain boundary.


To further enhance the physical consistency of the reconstructions, we incorporate physics-based regularization directly over the spatiotemporal domain. This is achieved by encouraging the physics-based residuals derived from the governing PDEs to approach zero. However, because the dynamics evolve on the 3D manifold $M$, conventional Euclidean differential operators cannot be directly applied to evaluate the PDE residuals. Instead, spatial derivatives are formulated intrinsically on the manifold using the Laplace--Beltrami operator $\Delta_M$ \citep{zhang2025physics}, which generalizes the Euclidean Laplacian to curved surfaces. Under the Neumann boundary conditions defined in Eq.~\eqref{eq:pdes}, the Laplace--Beltrami formulation is equivalent to the restriction of the Euclidean Laplacian to the manifold surface.
Then, the physics-based residuals are defined as:
\begin{equation}
    \begin{aligned}
        \mathcal{R}_u(\cdot) &:= \frac{\partial \hat{u}}{\partial t} - e_1 \Delta_M \hat{u} - g_1(\hat{u},\hat{v}), \\
        \mathcal{R}_v(\cdot) &:= \frac{\partial \hat{v}}{\partial t} - e_2 \Delta_M \hat{v} - g_2(\hat{u},\hat{v})
    \end{aligned}
    \label{eq:residuals}
\end{equation}
where $\hat{u}$ and $\hat{v}$ correspond to the channel components extracted from prediction $\widehat{\mathbf{Q}}_h$.




The total physics loss aggregates the PDE residuals across the temporal sequence:
\begin{equation}
    \begin{aligned}
        \mathcal{L}_{\mathrm{phy}} = &\sum_{n=0}^{B-1} \Bigl( \bigl\| \mathcal{R}_u(\widehat{\mathbf{Q}}_h(t+n)) \bigr\|_F^2 + \bigl\| \mathcal{R}_v(\widehat{\mathbf{Q}}_h(t+n)) \bigr\|_F^2 \Bigr) \\
        &+ \sum_{n=1}^{B-1} \Bigl( \bigl\| \mathcal{R}_u(\widehat{\mathbf{Q}}_h^K(t+n)) \bigr\|_F^2 + \bigl\| \mathcal{R}_v(\widehat{\mathbf{Q}}_h^K(t+n)) \bigr\|_F^2 \Bigr)
    \end{aligned}
    \label{eq:phys_loss}
\end{equation}
 By constraining both the static spatial reconstructions of the autoencoder and the temporal predictions of the Koopman operator, this dual-penalty mechanism ensures that the network produces physically plausible HR fields from sparse LR observations.


   \section{Theoretical Analysis of SR Error}
\label{s:error}

This section provides theoretical guarantees demonstrating how the proposed architecture inherently mitigates SR error through physics-informed constraints. 
Let $h^*: \mathbb{R}^{C \times N_\text{s}} \to \mathbb{R}^{C \times N_\text{s}^*}$ denote the true HR mapping, exactly satisfying the PDEs with the expected physics residual $\mathcal{R}(h^*) = 0$. For a given observation snapshot $\mathbf{Q}_l(t)$, its relationship to the true HR state is modeled as:
\begin{equation}
    \tilde{\mathbf{Q}}_l(t) = \mathbf{P}_{h \to l}^{\top} \, h^*(\mathbf{Q}_l(t)) + \mathbf{\Xi}_\text{mea}(t)
    \label{Eq: observation}
\end{equation}
 where $\mathbf{P}_{h \to l}$ maps the dense HR mesh onto the sparse LR space and $\mathbf{\Xi}_\text{mea}(t)$ represents the matrix of zero-mean measurement noise. Correspondingly, let $h: \mathbb{R}^{C \times N_\text{s}} \to \mathbb{R}^{C \times N_\text{s}^*}$ define the parameterized candidate mapping that attempts to reconstruct the HR field from the LR observations. The primary objective is to reconstruct an HR field $\widehat{\mathbf{Q}}_h(t) = h(\tilde{\mathbf{Q}}_l(t))$ such that the predicted state is both consistent with the empirical measurements and physically plausible. The corresponding reconstruction error is quantified directly in the HR space as
\begin{equation}
\mathcal{E}_{\text{HR}}(h) = \mathbb{E} \bigl[ \|h(\tilde{\mathbf{Q}}_l(t)) - h^*(\mathbf{Q}_l(t))\|_F^2 \bigr]
\label{Eq: e_hr}
\end{equation}
where $\|\cdot\|_F$ denotes the Frobenius norm evaluated over the dense HR mesh.


Let $\mathcal{H}$ denote the hypothesis space encompassing all mappings $h: \mathbb{R}^{C \times N_\text{s}} \to \mathbb{R}^{C \times N_\text{s}^*}$ by the unconstrained P-K-GCN (comprising encoder $\mathrm{E}_\theta$, Koopman matrix $\mathbf{K}$, and decoder $\mathrm{D}_\theta$). Given the high expressivity of over-parameterized neural networks, $\mathcal{H}$ possesses the capacity to fit unstructured noise. To mitigate overfitting, physical regularizers are introduced to constrain the optimization to a subset of mappings that adhere to the governing physics:

\begin{definition}[Physics-constrained hypothesis space]\label{def:physics_space}
For a given tolerance $\epsilon \ge 0$, the constrained space is defined as
$$\mathcal{H}_{\text{phy}}(\epsilon) = \left\{ h \in \mathcal{H} \;:\; \mathcal{R}(h) \le \epsilon \right\}$$
where $\mathcal{R}(h) = \mathbb{E} [\mathcal{L}_{\text{phy}}(h(\tilde{\mathbf{Q}}_l(t)))]$ is the expected PDE residual, with $\mathcal{L}_{\text{phy}}$ defined via $\mathcal{R}_u, \mathcal{R}_v$ in Eq.~\ref{eq:residuals}. The true mapping intrinsically satisfies $\mathcal{R}(h^*) = 0$.
\end{definition}

Enforcing this physical consistency explicitly limits the functional capacity of the model, establishing a strict inclusion property:

\begin{property}[Strict inclusion]\label{prop:strict_inclusion}
For sufficiently small $\epsilon > 0$, $\mathcal{H}_{\text{phy}}(\epsilon) \subset \mathcal{H}$, 
which implies there exists at least one unconstrained mapping $h \in \mathcal{H}$ such that $\mathcal{R}(h) > \epsilon$. 
\end{property}

While $\mathcal{H}_{\text{phy}}(\epsilon)$ restricts mappings in the HR domain, empirical training supervision operates exclusively in the LR observational space. As such, analyzing the model generalization behavior requires defining the induced hypothesis classes within the LR space:

\begin{definition}[Induced LR hypothesis class]\label{def:induced_lr_space}
The induced class in the LR space is defined as:
$$\mathcal{H}_l = \left\{ h_l(\tilde{\mathbf{Q}}_l(t)) = \mathbf{P}_{h \to l}^{\top} h(\tilde{\mathbf{Q}}_l(t)) \;:\; h \in \mathcal{H} \right\}, ~~~~\mathcal{H}_{l,\text{phy}}(\epsilon) = \left\{ \mathbf{P}_{h \to l}^{\top} h \;:\; h \in \mathcal{H}_{\text{phy}}(\epsilon) \right\}$$
It follows inherently that $\mathcal{H}_{l,\text{phy}}(\epsilon) \subseteq \mathcal{H}_l$.
\end{definition}

\subsection{Rademacher Complexity and Generalization Bound}

To quantify the expressiveness of the hypothesis spaces, we employ the empirical Rademacher complexity~\citep{bartlett2002rademacher, mohri2018foundations}, which measures the capacity of a function class to fit random noise: 

\begin{definition}[Empirical Rademacher complexity]\label{def:rademacher_complexity}
 For the full sequence of temporal observations over time points $\mathcal{T}$ with $N_\text{t}$ steps, represented by the LR feature tensor $\tilde{\mathcal{Q}}_l = \{\tilde{\mathbf{Q}}_l(t_i)\}_{i=1}^{N_\text{t}}$ and a data-driven loss function $\mathcal{L}_{\mathrm{d}}$ bounded by a constant $c>0$, the empirical Rademacher complexity is defined as
$$\hat{\mathfrak{R}}_{\tilde{\mathcal{Q}}_l}(\mathcal{H}_l) = \mathbb{E}_{\boldsymbol{\sigma}} \left[ \sup_{h_l \in \mathcal{H}_l} \frac{1}{N_\text{t}} \sum_{i=1}^{N_\text{t}} \sigma_i \, \mathcal{L}_{\mathrm{d}}\bigl(h_l(\tilde{\mathbf{Q}}_l(t_i)),\, \tilde{\mathbf{Q}}_l(t_i)\bigr) \right]$$
where $\sigma_i$ are i.i.d. Rademacher variables taking values $\pm1$ with equal probability. 
\end{definition}


By restricting the hypothesis space to physically plausible mappings (Definition~\ref{def:physics_space}) , the model capacity to correlate with unstructured noise is inherently limited, as formalized in the following proposition:


\begin{proposition}[Complexity reduction via physics constraint]\label{prop:complexity_reduction}
Assume that $\mathcal{L}_{\mathrm{d}}$ is bounded and there exists at least one Rademacher sequence $\boldsymbol{\sigma} \in \{-1,+1\}^{N_\text{t}}$ for which the unconstrained supremum
\[
\sup_{h \in \mathcal{H}_l} \frac{1}{N_\text{t}} \sum_{i=1}^{N_\text{t}} \sigma_i \mathcal{L}_{\mathrm{d}}(h(\tilde{\mathbf{Q}}_l(t_i)), \tilde{\mathbf{Q}}_l(t_i))
\]
is achieved by some function $h^* \notin \mathcal{H}_{l,\mathrm{phy}}(\epsilon)$. Then for sufficiently small $\epsilon > 0$,
\[
\hat{\mathfrak{R}}_{\tilde{\mathcal{Q}}_l}(\mathcal{H}_{l,\mathrm{phy}}(\epsilon)) < \hat{\mathfrak{R}}_{\tilde{\mathcal{Q}}_l}(\mathcal{H}_l)
\]
\end{proposition}


To establish the aforementioned proposition, observe that for any fixed Rademacher sequence $\boldsymbol{\sigma}$, the set inclusion $\mathcal{H}_{l,\mathrm{phy}}(\epsilon) \subseteq \mathcal{H}_l$ yields
\begin{equation}
\sup_{h \in \mathcal{H}_{l,\mathrm{phy}}(\epsilon)} \frac{1}{N_\text{t}} \sum_{i=1}^{N_\text{t}} \sigma_i \mathcal{L}_{\mathrm{d}}(h(\tilde{\mathbf{Q}}_l(t_i)), \tilde{\mathbf{Q}}_l(t_i)) \le \sup_{h \in \mathcal{H}_l} \frac{1}{N_\text{t}} \sum_{i=1}^{N_\text{t}} \sigma_i \mathcal{L}_{\mathrm{d}}(h(\tilde{\mathbf{Q}}_l(t_i)), \tilde{\mathbf{Q}}_l(t_i))
\end{equation}
Taking expectations over $\boldsymbol{\sigma}$ preserves the inequality, yielding $\hat{\mathfrak{R}}_{\tilde{\mathcal{Q}}_l}(\mathcal{H}_{l,\mathrm{phy}}(\epsilon)) \le \hat{\mathfrak{R}}_{\tilde{\mathcal{Q}}_l}(\mathcal{H}_l)$. To establish strictness, observe that by assumption there exists a set $S \subseteq \{-1,+1\}^{N_\text{t}}$ of positive probability such that for every $\boldsymbol{\sigma} \in S$, the unconstrained supremum is attained by some $h^* \notin \mathcal{H}_{l,\mathrm{phy}}(\epsilon)$. For these sequences, the inequality above is strict, leading to $\hat{\mathfrak{R}}_{\tilde{\mathcal{Q}}_l}(\mathcal{H}_{l,\mathrm{phy}}(\epsilon)) < \hat{\mathfrak{R}}_{\tilde{\mathcal{Q}}_l}(\mathcal{H}_l)$. 
Following standard statistical learning theory, for each induced LR mapping $h_l \in \mathcal{H}_l$, the expected data-driven risk $\mathcal{E}_{\text{LR}}(h_l)$ is bounded by its empirical counterpart $\hat{\mathcal{E}}_{\text{LR}}(h_l)$ and the empirical Rademacher complexity $\hat{\mathfrak{R}}_{\tilde{\mathcal{Q}}_l}(\mathcal{H}_l)$:

\begin{theorem}[Uniform convergence bound \citep{mohri2018foundations} ]\label{thm:uniform_convergence}
For any $\delta \in (0,1)$, with probability at least $1-\delta$ over the draw of $\tilde{\mathcal{Q}}_l$, every $h_l \in \mathcal{H}_l$ satisfies
$$\mathcal{E}_{\text{LR}}(h_l) \le \hat{\mathcal{E}}_{\text{LR}} + 2\hat{\mathfrak{R}}_{\tilde{\mathcal{Q}}_l}(\mathcal{H}_l) + 3c\sqrt{\frac{\ln(2/\delta)}{2N_\text{t}}}$$
where $\mathcal{E}_{\text{LR}}(h_l) = \mathbb{E}_{\tilde{\mathbf{Q}}_l(t)}[\mathcal{L}_{\mathrm{d}}(h_l(\tilde{\mathbf{Q}}_l(t)), \tilde{\mathbf{Q}}_l(t))]$.
\end{theorem}


Minimizing the LR risk (Theorem~\ref{thm:uniform_convergence}) does not deterministically guarantee a small HR reconstruction error because the linear projection $\mathbf{P}_{h \to l}$ is not injective. To address this inherent ill-posedness issue, we leverage the regularizing effect of the physics constraint, which enforces conditional stability on the inverse SR problem.

\begin{assumption}[Conditional stability]\label{assum:conditional_stability}
There exists a constant $C_{\text{stab}} > 0$ such that for any $h_1, h_2 \in \mathcal{H}_{\text{phy}}(\epsilon)$,
$$\|h_1 - h_2\|_F \le C_{\text{stab}} \cdot \|\mathbf{P}_{h \to l}^{\top}(h_1 - h_2)\|_F$$
\end{assumption}


Assumption \ref{assum:conditional_stability} establishes a standard well-posedness condition for PDE-constrained inverse problems. It enables the HR reconstruction error to be strictly bounded by the observable LR error as given in Lemma \ref{lem:stability_bound}.

\begin{lemma}[Stability bound]\label{lem:stability_bound}
According to Assumption~\ref{assum:conditional_stability}, the reconstruction error satisfies:
$$\mathcal{E}_{\text{HR}}(h) \le C_{\text{stab}}^2 \, \mathcal{E}_{\text{LR}}(\mathbf{P}_{h \to l}^{\top} h) + \mathcal{O}(\epsilon)$$
\end{lemma}


Applying Assumption~\ref{assum:conditional_stability} to the candidate mapping $h$ and the true mapping $h^*$ for an observation $\tilde{\mathbf{Q}}_l(t)$ yields:
\begin{equation}
\|h(\tilde{\mathbf{Q}}_l(t)) - h^*(\mathbf{Q}_l(t))\|_F \le C_{\text{stab}} \|\mathbf{P}_{h \to l}^{\top}(h(\tilde{\mathbf{Q}}_l(t)) - h^*(\mathbf{Q}_l(t)))\|_F
\label{lemma1-1}
\end{equation}
Applying the definition of $\mathcal{E}_{\text{HR}}(h)$ in Eq. (\ref{Eq: e_hr}) yields
\begin{equation}
\mathcal{E}_{\text{HR}}(h) \le C_{\text{stab}}^2 \, \mathbb{E} \left[ \|\mathbf{P}_{h \to l}^{\top}(h(\tilde{\mathbf{Q}}_l(t)) - h^*(\mathbf{Q}_l(t)))\|_F^2 \right]
\label{lemma1-2}
\end{equation}

Under the observation model in Eq. (\ref{Eq: observation}), the independence of measurement noise $\mathbf{\Xi}_\text{mea}(t)$ yields: 
\begin{equation}
\mathbb{E} \left[ \|\mathbf{P}_{h \to l}^{\top} h(\tilde{\mathbf{Q}}_l(t)) - \tilde{\mathbf{Q}}_l(t)\|_F^2 \right] = \mathbb{E} \left[ \|\mathbf{P}_{h \to l}^{\top}(h(\tilde{\mathbf{Q}}_l(t))-h^*(\mathbf{Q}_l(t)))\|_F^2 \right] + \sigma_\text{mea}^2
\label{lemma1-3}
\end{equation}
where $\sigma_\text{mea}^2 = \mathbb{E}[\|\mathbf{\Xi}_\text{mea}(t)\|_F^2]$. Rearranging this relationship isolates the projected structural error, which can then be bounded by the expected LR loss, denoted as $\mathcal{E}_{\text{LR}}$:
\begin{equation}
\mathbb{E}\Bigl[ \|\mathbf{P}_{h\to l}^{\top}\bigl(h(\tilde{\mathbf{Q}}_l(t))-h^*(\mathbf{Q}_l(t))\bigr)\|_F^2 \Bigr]
\leq \mathcal{E}_{\mathrm{LR}}(\mathbf{P}_{h\to l}^{\top} h)
\label{lemma1-4}
\end{equation}
where $\mathcal{E}_{\mathrm{LR}}(\mathbf{P}_{h\to l}^{\top} h) = \mathbb{E}[\|\mathbf{P}_{h\to l}^{\top} h(\tilde{\mathbf{Q}}_l(t)) - \tilde{\mathbf{Q}}_l(t)\|_F^2]$.



A more detailed expansion of the LR loss reveals the presence of a coupling term $\mathbb{E}\bigl[\langle \mathbf{P}_{h\to l}^{\top}(h-h^*),\,\mathbf{\Xi}_\text{mea}\rangle_F\bigr]$ that does not vanish because $h$ depends on the input $\tilde{\mathbf{Q}}_l$. The incorporation of physics-based constraints (i.e., $\mathcal{R}(h) \le \epsilon$) mitigates this effect by restricting the admissible hypothesis space to physically consistent solutions. In particular, under the well-posedness assumptions of the governing PDE system, the physics residual $\mathcal{R}(h)$ provides control over the deviation between the reconstructed solution and the true physical dynamics.
Specifically, there exists a constant $C_{\mathcal{R}}$ such that for every $h\in\mathcal{H}_{\text{phy}}(\epsilon)$,
\begin{equation}
\|h(\tilde{\mathbf{Q}}_l)-h^*(\mathbf{Q}_l)\|_F \le C_{\mathcal{R}}\,\epsilon \sim\mathcal{O}(\epsilon) \qquad\text{almost surely}
\end{equation}
Using this uniform bound and the submultiplicativity of the Frobenius norm, we estimate
\begin{equation}
\begin{aligned}
\bigl|\mathbb{E}\bigl[\langle \mathbf{P}_{h\to l}^{\top}(h-h^*),\,\mathbf{\Xi}_\text{mea}\rangle_F\bigr]\bigr|
&\le \mathbb{E}\bigl[\|\mathbf{P}_{h\to l}^{\top}(h-h^*)\|_F\,\|\mathbf{\Xi}_\text{mea}\|_F\bigr] \\
&\le \|\mathbf{P}_{h\to l}^{\top}\|_{\mathrm{op}}\; \mathbb{E}\bigl[\|h-h^*\|_F\,\|\mathbf{\Xi}_\text{mea}\|_F\bigr]\sim \mathcal{O}(\epsilon)
\end{aligned}
\end{equation}
where $\|\cdot\|_{\mathrm{op}}$ denotes the operator norm (spectral norm) of the projection matrix.

Inserting this bound into the following exact decomposition :
\begin{equation}
\mathcal{E}_{\text{LR}}(\mathbf{P}_{h\to l}^{\top}h)
= \mathbb{E}\bigl[\|\mathbf{P}_{h\to l}^{\top}(h-h^*)\|_F^2\bigr] + \sigma_{\text{mea}}^2 - 2\mathbb{E}\bigl[\langle \mathbf{P}_{h\to l}^{\top}(h-h^*),\,\mathbf{\Xi}_\text{mea}\rangle_F\bigr]
\end{equation}
we obtain
\begin{equation}
\mathbb{E}\bigl[\|\mathbf{P}_{h\to l}^{\top}(h-h^*)\|_F^2\bigr]
= \mathcal{E}_{\text{LR}}(\mathbf{P}_{h\to l}^{\top}h) - \sigma_{\text{mea}}^2 + \mathcal{O}(\epsilon)
\le \mathcal{E}_{\text{LR}}(\mathbf{P}_{h\to l}^{\top}h) + \mathcal{O}(\epsilon)
\label{lemma1-6}
\end{equation}
Substituting Eq.~\eqref{lemma1-6} into Eq.~\eqref{lemma1-2} yields the theoretical bound in Lemma~\ref{lem:stability_bound}.


\subsection{SR Error Bounds under Physics Augmentation}

Integrating the LR generalization bound (Theorem~\ref{thm:uniform_convergence}) and the stability lemma (Lemma~\ref{lem:stability_bound}) yields the primary theoretical guarantee of the proposed framework:

\begin{theorem}[Generalization bound for SR error]\label{thm:generalization_bound}
Under Assumption~\ref{assum:conditional_stability}, with probability at least $1-\delta$ over the draw of $\tilde{\mathcal{Q}}_l$, for every $h \in \mathcal{H}_{\text{phy}}(\epsilon)$, we have
$$\mathcal{E}_{\text{HR}}(h) \le C_{\text{stab}}^2 \left( \hat{\mathcal{E}}_{\text{LR}}(\mathbf{P}_{h \to l}^{\top} h) + 2\hat{\mathfrak{R}}_{\tilde{\mathcal{Q}}_l}(\mathcal{H}_{l,\text{phy}}(\epsilon)) + 3c\sqrt{\frac{\ln(2/\delta)}{2N_\text{t}}} \right) + \mathcal{O}(\epsilon)$$
\end{theorem}

Applying Theorem~\ref{thm:uniform_convergence} to the induced LR mapping $h_l = \mathbf{P}_{h \to l}^{\top} h \in \mathcal{H}_{l,\mathrm{phy}}(\epsilon)$ bounds the expected LR risk $\mathcal{E}_{\mathrm{LR}}(\mathbf{P}_{h \to l}^{\top} h)$:
\begin{equation}
\mathcal{E}_{\text{LR}}(\mathbf{P}_{h \to l}^{\top} h) \le \hat{\mathcal{E}}_{\text{LR}}(\mathbf{P}_{h \to l}^{\top} h) + 2\hat{\mathfrak{R}}_{\tilde{\mathcal{Q}}_l}(\mathcal{H}_{l,\text{phy}}(\epsilon)) + 3c\sqrt{\frac{\ln(2/\delta)}{2N_\text{t}}}
\end{equation}
Substituting this result into the Stability Bound (Lemma~\ref{lem:stability_bound}) concludes the proof.

Theorem~\ref{thm:generalization_bound} provides the definitive theoretical evidence explaining how physics augmentation mitigates SR error. By restricting the hypothesis space to mappings that adhere to the governing dynamics (Definition~\ref{def:physics_space}), the model capacity to overfit spurious noise is strictly reduced, which mathematically manifests as a lower Rademacher complexity (Proposition~\ref{prop:complexity_reduction}). The diminished complexity tightens the generalization bound in the LR observable space (Theorem~\ref{thm:uniform_convergence}). Finally, through the conditional stability of the inverse problem (Assumption~\ref{assum:conditional_stability}), Lemma~\ref{lem:stability_bound} successfully translates the tightened LR boundary into a strictly bounded HR reconstruction error. Therefore, Theorem~\ref{thm:generalization_bound} mathematically guarantees that even if a purely data-driven network and the physics-augmented P-K-GCN achieve identical empirical training losses ($\hat{\mathcal{E}}_{\text{LR}}$), the P-K-GCN will inherently yield superior and more robust SR performance due to the physics-driven reduction in the error bound.
Similarly, Supplementary Material \ref{subsec:koopman_theory} provides theoretical analysis of error mitigation via Koopman regularization.

  \section{Experimental Design and Results} \label{s:results}

\subsection{Model Architecture}

\begin{figure}[!ht]
	\begin{center}
		\includegraphics[width=5.5in]{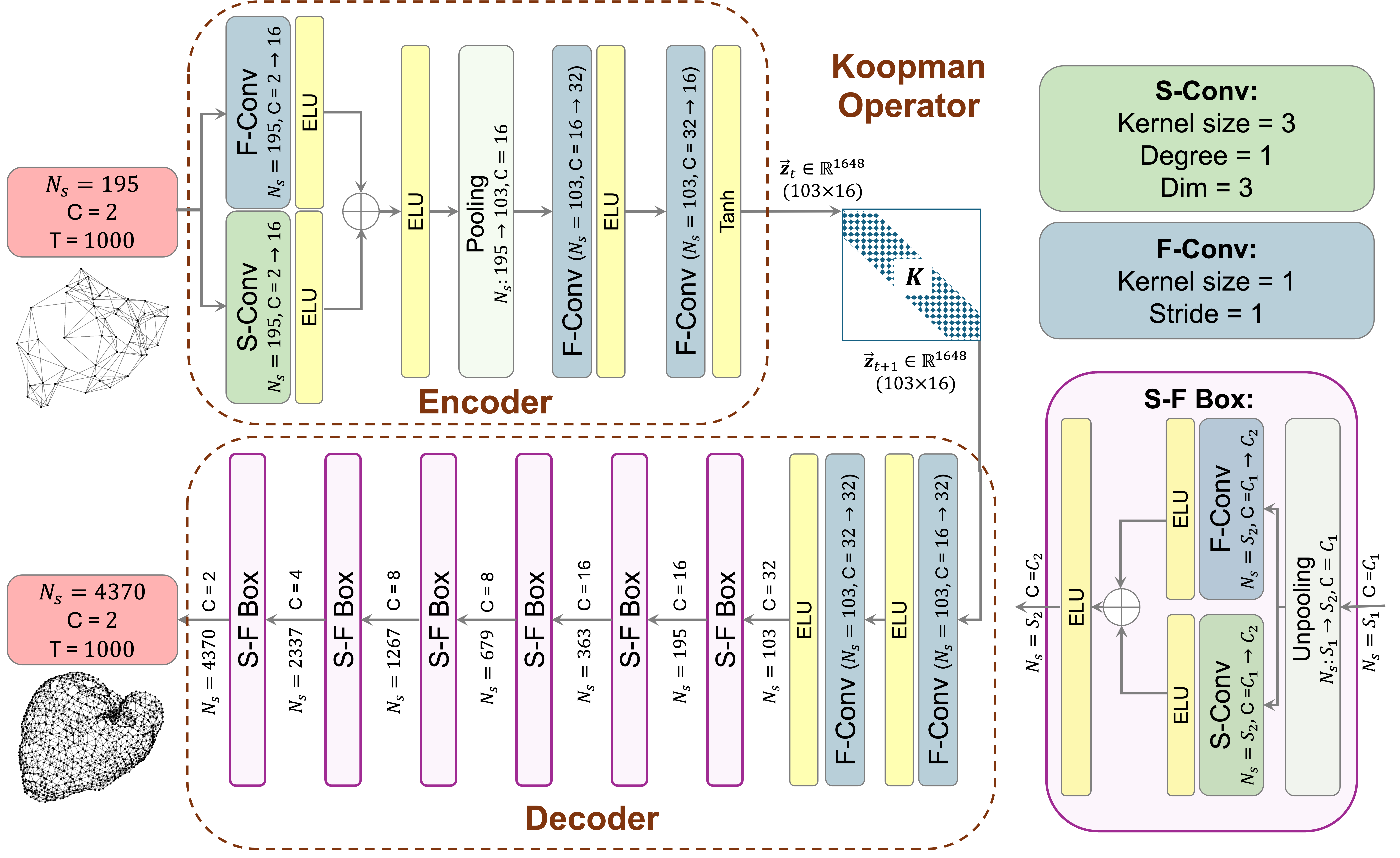}
		\caption{The architecture detail of the proposed P-K-GCN framework}
		\label{Fig:NN_details}
	\end{center}    
\end{figure}

Fig. \ref{Fig:NN_details} illustrates the P-K-GCN architecture, which processes spatiotemporal inputs on a coarse irregular mesh of 195 nodes with 2 channels and 1000 time steps. The encoder extracts spatial features using a combination of S-Conv and F-Conv layers alongside a pooling layer to generate a compact, 1648-dimensional latent vector. Furthermore, a learnable linear operator $\mathbf{K}$ systematically propagates the latent state forward. The predicted latent state is then fed into the decoder to reconstruct the HR field. Using F-Conv layers followed by a sequence of 6 S-F blocks, which combine unpooling with residual convolutions, the decoder progressively achieves SR of the LR input. This hierarchical upsampling produces a final HR output of 4370 spatial nodes. 

\subsection{Data Preparation}
We assess the performance of our P-K-GCN framework in reconstructing HR cardiac electrodynamics within a 3D ventricular geometry from LR observations. The anatomical domain is discretized into 4,370 nodes and 8,736 mesh elements, forming a refined computational mesh derived from the geometry dataset provided in the 2007 PhysioNet Computing in Cardiology Challenge \cite{goldberger2000physiobank}. We simulate the propagation of cardiac electrical activity by numerically solving the Aliev-Panfilov (AP) model with the reaction functions defined as follows:
\begin{equation}
      \begin{aligned}
      g_1(u,v) &= C_1 u (u - \alpha)(1 - u) - C_2 u v,  \\
      g_2(u,v) &=\xi(u, v)\left( -v - C_1 u (u - \alpha - 1) \right),\\
      \xi(u, v) &= e_0 + (\mu_1 v)/(u + \mu_2)
  \end{aligned}
\end{equation}
where $u$ is the normalized transmembrane potential and $v$ is the recovery dynamics of the electrical excitation. Adopted from Ref.~\cite{aliev1996simple}, the AP model parameters are set to: $\alpha=0.1$, $C_1=8$, $C_2=1$, $e_0=0.002$, and $\mu_1=\mu_2=0.3$; and diffusion parameters set to: $e_1=6$ and $e_2=0$. An additional ectopic activation source is introduced in the right ventricle, spatially separated from the primary pacing site, to induce self-sustained and spatially discordant propagation patterns that mimic the complex fibrillatory dynamics observed in cardiac arrhythmias. We denote the resulting simulation data as $\boldsymbol{q}(\boldsymbol{x}, t)=[u(\boldsymbol{x}, t),v(\boldsymbol{x}, t)]=[u(\boldsymbol{x}_i, t_j),v(\boldsymbol{x}_i, t_j)]_{i\in \mathcal{X}_{h/l} ,j\in \mathcal{T}}$ with $|\mathcal{X}_h|=4370, |\mathcal{X}_l|=195$, and $|\mathcal{T}|=1000$.
Because measurement noise is inevitable in real-world data collection, we add different levels of noise to the simulation data to investigate prediction performance. Specifically, the physical measurements are generated as $\tilde{\boldsymbol{q}}(\boldsymbol{x}, t) = \boldsymbol{q}(\boldsymbol{x}, t) + \boldsymbol{\xi}(\boldsymbol{x}, t)$, where $\boldsymbol{\xi}(\mathbf{x}, t) \sim \sigma_\xi \cdot \mathcal{N}(\mathbf{0}, 1)$ denotes the noise term (which corresponds to the matrix noise $\mathbf{\Xi}_\text{mea}(t)$), with $\sigma_\xi$ representing the noise level coefficient.


We benchmark P-K-GCN against three widely adopted methods: \textbf{(i) Traditional Neural Network (NN)} -- A purely data-driven, standard deep learning architecture lacking geometry-aware spatial modeling, stable temporal modeling, and physical constraints; \textbf{(ii) K-GCN} -- An ablated version of our framework that utilizes geometry-aware spatial modeling and Koopman-enhanced temporal modeling, but omits the physics-based regularization; \textbf{(iii) PINN} -- A conventional physics-informed approach that encodes the AP equations directly into the loss function, but lacks both geometry-aware spatial mapping and stable temporal modeling. 
We employ the relative error ($RE$) to assess the model performance:
\begin{equation}
    RE = \frac{\| \hat{\boldsymbol{q}}(\boldsymbol{x}, t) - \boldsymbol{q}(\boldsymbol{x}, t) \|}{ \| \boldsymbol{q}(\boldsymbol{x}, t) \|}
    \label{eq:relative_error}
\end{equation}
where $\boldsymbol{q}(\boldsymbol{x}, t)$ and $\hat{\boldsymbol{q}}(\boldsymbol{x}, t)$ denote the reference and reconstructed dynamics, respectively.

\subsection{Experiment Results Analysis}

\begin{figure}[!ht]
	\begin{center}
		\includegraphics[width=4.8in]{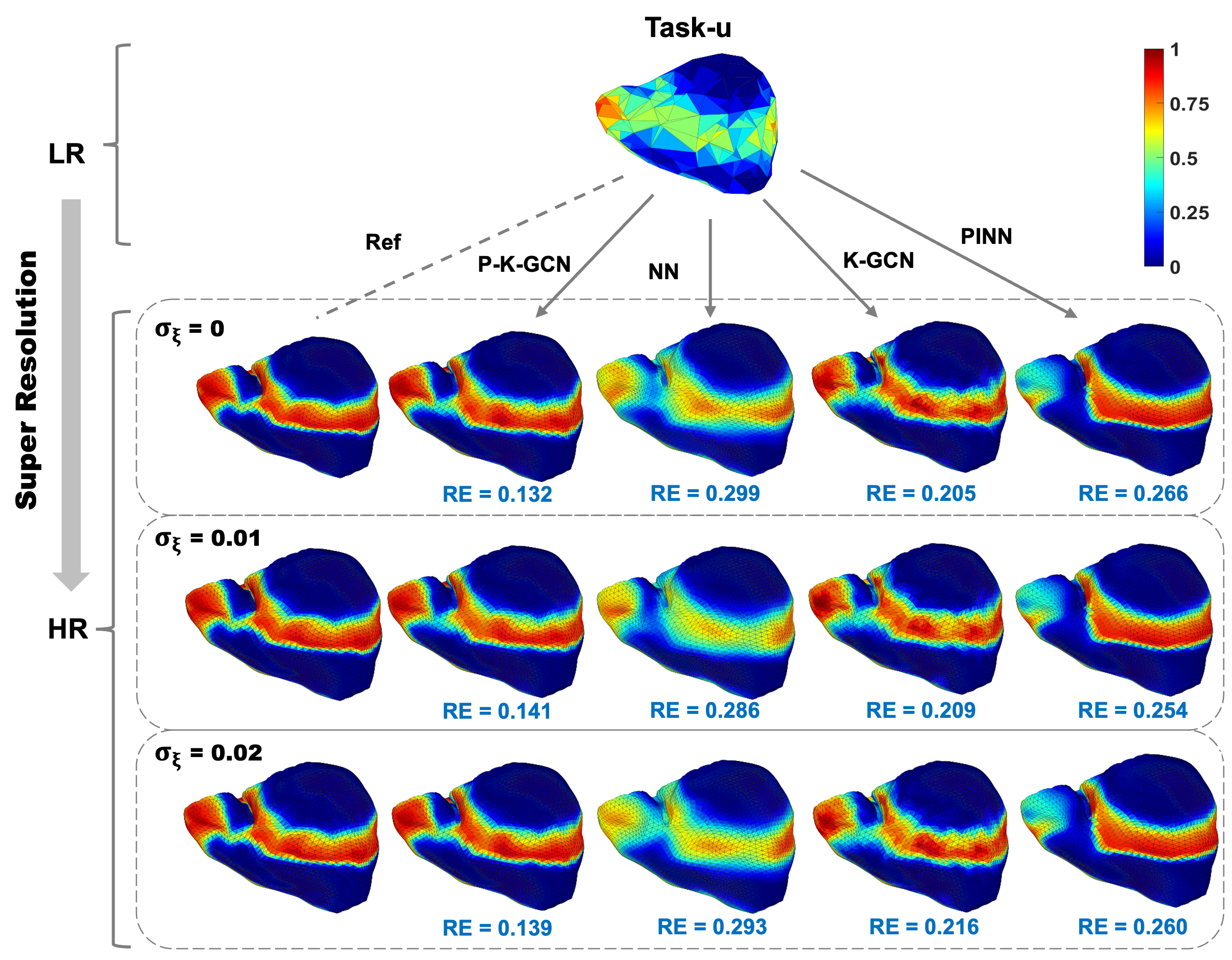}
		\caption{Visual comparison of the reconstructed transmembrane potential ($u$) at time step 35 under varying noise levels ($\sigma_\xi = 0, 0.01, 0.02$).
        }
		\label{Fig:Reconstruction-u}
	\end{center}    
\end{figure}

\begin{figure}[!ht]
	\begin{center}
		\includegraphics[width=4.8in]{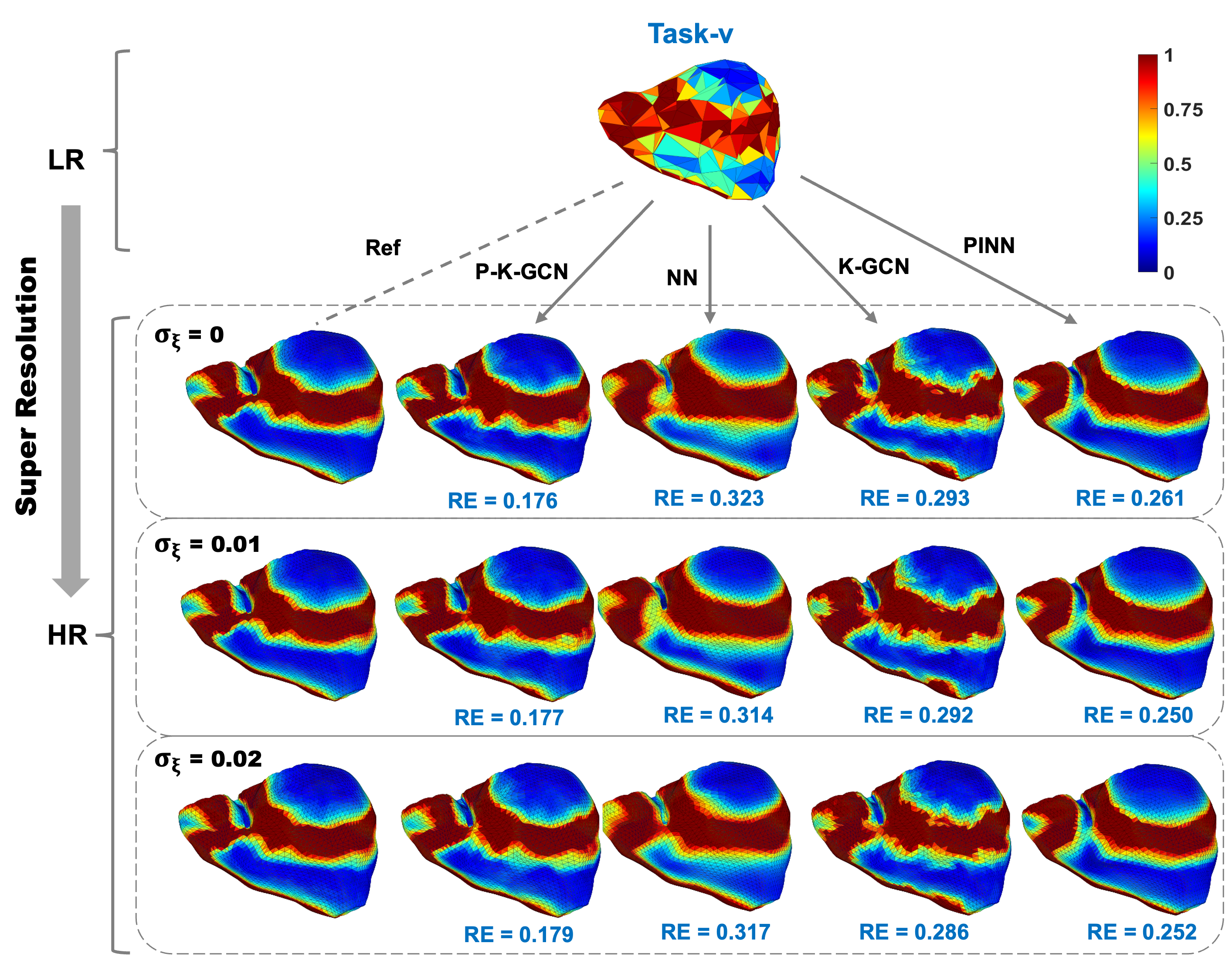}
		\caption{Visual comparison of the reconstructed recovery variable ($v$) at time step 35 under varying noise levels ($\sigma_\xi = 0, 0.01, 0.02$). 
        }
		\label{Fig:Reconstruction-v}
	\end{center}    
\end{figure}

Fig.~\ref{Fig:Reconstruction-u} presents a visual comparison of the reconstructed transmembrane potential $u$ at time step 35 (out of 1,000) across different methods and noise levels. 
From the visual comparison, it is evident that our P-K-GCN achieves the best reconstruction accuracy, maintaining high fidelity to the ground truth across all noise conditions. This superior performance is achieved by integrating geometry-aware spatial graph modeling and Koopman-enhanced stable temporal modeling with physics-augmented regularization. 
In contrast, the standard NN performs the worst, consistently failing to recover fine-scale structural details of the electrodynamics. 
The PINN predictions capture the general wave patterns but yield overly smooth reconstructions that blur critical local features. 
The K-GCN successfully predicts the majority of the spatial features by utilizing its graph-based architecture and Koopman operator to handle irregular geometries and stabilize temporal evolution. However, it exhibits significant visual artifacts across all noise levels because of the absence of physics-based constraints, leading to unstable dynamic evolution when interpolating from sparse or noisy inputs. It is worth noting that similar performance is observed for the recovery variable ($v$ signal), as shown in Fig. \ref{Fig:Reconstruction-v}.

\begin{figure}[!ht]
	\begin{center}
		\includegraphics[width=5in]{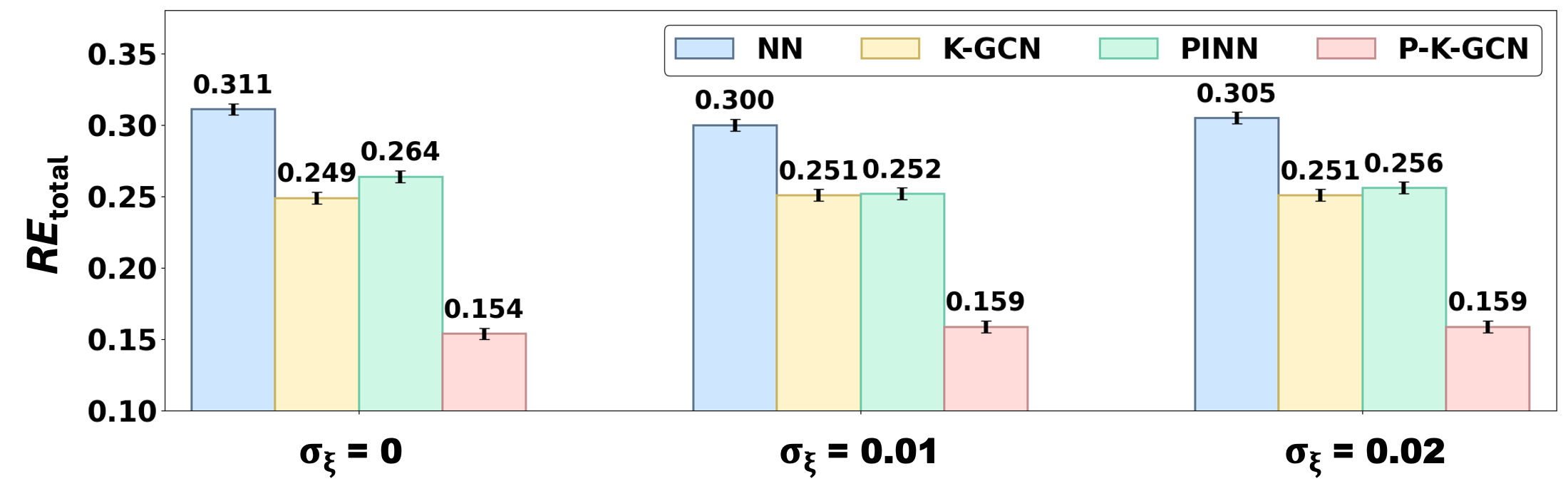}
		\caption{Bar chart comparing $RE_{\text{total}}$ of our P-K-GCN framework against benchmark methods (NN, K-GCN, PINN) under varying noise levels ($\sigma_\xi$ = 0, 0.01, 0.02).}
		\label{Fig:RE}
	\end{center}    
\end{figure}

Fig.~\ref{Fig:RE} compares the quantitative prediction performance of P-K-GCN with NN, K-GCN, and PINN under three noise levels ($\sigma_\xi$ = 0, 0.01, 0.02), based on the aggregated relative error metric ($RE_{\text{total}} := \frac{1}{2}(RE_u+RE_v)$) derived from triplicate experiments with randomized seeds. When $\sigma_\xi=0$, our P-K-GCN establishes a strong baseline with an $RE_{\text{total}}$ of $0.154 \pm (1.5 \times 10^{-4})$. This represents a substantial accuracy improvement, achieving error reductions of $50.48\%$, $38.15\%$, and $41.67\%$ over NN ($0.311 \pm (4.0 \times 10^{-3})$), K-GCN ($0.249 \pm (7.4 \times 10^{-4})$), and PINN ($0.264 \pm (3.7 \times 10^{-3})$), respectively. 
As the noise level escalates from $0.01$ to $0.02$, the predictive error of P-K-GCN remains remarkably stable at $0.159$ (with minimal standard deviations of $\pm (2.8 \times 10^{-5})$ and $\pm (3.8 \times 10^{-4})$, respectively). P-K-GCN consistently maintains substantial performance margins, outperforming the baseline models by approximately 47\% (NN), 36\% (K-GCN), and 37\% (PINN) across both noise levels. These results suggest that the integration of Koopman-based latent dynamics modeling and physics-constrained regularization effectively suppresses noise amplification and stabilizes the reconstruction process under corrupted measurement conditions.



\section{Conclusions} \label{s:conclusions}

This paper presents a novel Physics-augmented Koopman-enhanced Graph Convolutional Network (P-K-GCN) for the spatiotemporal super-resolution of complex dynamic systems on irregular domains. By integrating geometry-aware spatial modeling and Koopman-based temporal dynamics with physics-informed regularization, our framework addresses the fundamental limitations of purely data-driven approaches in handling high-dimensional, nonlinear systems.
Additionally, we mathematically demonstrate that the imposition of physical priors and Koopman temporal regularizers guarantees the mitigation of spatial SR error and prevents the exponential accumulation of temporal errors.
We validated the P-K-GCN on reconstruction high-resolution 3D cardiac electrodynamics. Numerical experiments demonstrate that our method achieves superior reconstruction accuracy and noise resilience compared to existing methods commonly used in current practice. Owing to its ability to handle irregular geometries, sparse observations, and complex nonlinear dynamics, our P-K-GCN framework has broad potential applicability for reliable high-resolution reconstruction across a wide range of scientific, engineering, and biomedical spatiotemporal systems.

  \section{Data Availability Statement}
    The mesh and geometry data supporting the findings of this study are openly available in the PhysioNet/Computing in Cardiology Challenge 2007 at \url{https://physionet.org/content/challenge-2007/1.0.0/}, reference number ~\cite{goldberger2000physiobank}. 
    

  \section*{Acknowledgement}
  This research work was supported by the National Heart, Lung, And Blood Institute of the National Institutes of Health under Award Number R01HL172292. The content is solely the responsibility of the authors and does not necessarily represent the official views of the National Institutes of Health. 

  {
    \bibliographystyle{apalike}
    \spacingset{1}
    \bibliography{Ref}
  }

  \appendix
\begin{center}
    {\Huge \textbf{Supplementary Material}} \vspace{1em} \\
\end{center}
\vspace{1em}
This supplementary document provides additional details, figures, and results that support the main manuscript titled: "\textit{P-K-GCN: Physics-augmented Koopman-enhanced Graph Convolutional Network for Deep Spatiotemporal Super-resolution}".

\section{Theoretical Analysis of Error Mitigation via Koopman Regularization}
\label{subsec:koopman_theory}


   In standard dynamic modeling, the true nonlinear temporal evolution $\mathbf{Q}(t+n) = \mathcal{F}^n(\mathbf{Q}(t))$ is typically approximated by a learned nonlinear transition function $\mathcal{M}_\theta$. If $\mathcal{M}_\theta$ possesses a local Lipschitz constant $L_{\mathcal{M}} > 1$, any initial spatial reconstruction error or measurement noise $\epsilon(t)$ compounds exponentially over time, scaling as $\mathcal{O}(L_{\mathcal{M}}^n \epsilon(t))$ (see Supplementary Material~\ref{app:normal-error} for detailed derivation). 
To mitigate this explosive divergence, the P-K-GCN framework restricts the temporal transition to an approximately linear evolution within an optimized latent observable space, subject to explicit norm regularization.

\subsection{Koopman-Constrained Temporal Hypothesis Space and Error Accumulation}

The unconstrained temporal hypothesis space, $\mathcal{H}_{\mathrm{temp}}$, which encompasses arbitrary nonlinear recurrent mappings, is highly expressive but prone to overfitting temporal noise, leading to degraded long-term forecasting. To theoretically guarantee improved stability over an $n$-step inference horizon, we define a constrained subset of temporal mappings governed exclusively by the linear Koopman operator:

\begin{definition}[Koopman-constrained temporal space]\label{def:koopman_space}
The temporal hypothesis space constrained by the Koopman operation is defined as:
$$
    \mathcal{H}_{\text{Koop}} = \left\{ h_{\mathrm{temp}} \;:\; \widehat{\mathbf{Q}}_h^K(t+n) = \mathrm{D}_\theta\bigl(\widehat{\mathbf{Z}}^K(t+n)\bigr), \text{ where } \widehat{\mathbf{Z}}^K(t+n) = \operatorname{vec}^{-1}(\mathbf{K}^n \boldsymbol{z}(t)) \right\}
$$
where $\boldsymbol{z}(t) = \operatorname{vec}(\mathrm{E}_\theta(\tilde{\mathbf{Q}}_l(t)))$ denotes the vectorized encoded observable state prior to the application of the Koopman operator.
\end{definition}

In practical optimization, the trainable matrix $\mathbf{K}$ is subjected to standard weight regularization, which actively restricts its Frobenius norm, $\|\mathbf{K}\|_F$. Because the spectral norm is strictly bounded by the Frobenius norm ($\|\mathbf{K}\|_2 \le \|\mathbf{K}\|_F$), this structural prior explicitly restricts the operator's expansiveness, leading to significantly attenuated error propagation rate compared to unconstrained nonlinear sequence models:

\begin{proposition}[Attenuated Temporal Error Accumulation]\label{prop:attenuated_temporal_error} 
Assume that the Koopman matrix $\mathbf{K}$ is regularized during training with $\|\mathbf{K}\|_F \le \gamma$, where $\gamma < L_{\mathcal{M}}$. Because $\|\mathbf{K}\|_2 \le \|\mathbf{K}\|_F$, the temporal transition is bounded by a maximal expansion rate $\gamma$. For any initial latent representation error $e_z(t) = \|\boldsymbol{z}(t) - \boldsymbol{z}^*(t)\|_2$ at time $t$, and accumulated latent approximation error $\boldsymbol{\eta}_n$, the error at inference step $t+n$ is bounded by:
$$
    \|\boldsymbol{z}^K(t+n) - \boldsymbol{z}^*(t+n)\|_2 \le \|\mathbf{K}\|_2^n e_z(t) + \|\boldsymbol{\eta}_n\|_2 \le \gamma^n e_z(t) + \|\boldsymbol{\eta}_n\|_2
$$
\end{proposition}

The proof of Proposition \ref{prop:attenuated_temporal_error} is provided in Supplementary Material~\ref{app:n-error}. By explicitly isolating the approximation error $\boldsymbol{\eta}_n$, this bound demonstrates that the propagation of the initial measurement error $e_z(t)$ is controlled by $\mathcal{O}(\gamma^n)$. Because the regularization guarantees $\gamma \ll L_{\mathcal{M}}$, this represents a massive, systematic reduction in accumulated error compared to the $\mathcal{O}(L_{\mathcal{M}}^n \epsilon(t))$ explosion in unconstrained nonlinear models.

Furthermore, confining the trajectory space to linear transitions reduces the empirical Rademacher complexity of the model over the observation window as shown Lemma \ref{lem:temporal_complexity}: 

\begin{lemma}[Temporal Complexity Reduction]\label{lem:temporal_complexity}
Let $\hat{\mathfrak{R}}_{\tilde{\mathcal{Q}}_l}(\mathcal{H}_{\mathrm{temp}})$ denote the Rademacher complexity of standard recurrent transitions over a temporal block $\mathcal{B}_l$. By restricting the temporal progression to linear transitions, we achieve a strict complexity reduction:
$$
    \hat{\mathfrak{R}}_{\tilde{\mathcal{Q}}_l}(\mathcal{H}_{\text{Koop}}) < \hat{\mathfrak{R}}_{\tilde{\mathcal{Q}}_l}(\mathcal{H}_{\mathrm{temp}})
$$
\end{lemma}

  Under Frobenius norm regularization $\|\mathbf{K}\|_F \le \gamma$, the empirical Rademacher complexity of this linear hypothesis class over a temporal block $\mathcal{B}_l$ of size $B$ scales as $\mathcal{O}(\gamma / \sqrt{B})$ (up to logarithmic factors) for bounded latent inputs~\cite{bartlett2002rademacher}. This complexity is several orders of magnitude smaller than that typically associated with deep recurrent neural network architectures. This reduced complexity, together with the uniform convergence theorem (Theorem~\ref{thm:uniform_convergence}), guarantees that the LR error at the initial time step $\mathcal{E}_{\mathrm{LR}}^{(0)}$ tightly controls the expected reconstruction error, preventing overfitting (see Theorem~\ref{thm:koopman_generalization}). 

\subsection{SR Error Bounds under Koopman Theory}

Theorem~\ref{thm:koopman_generalization} establishes the SR error bounds under Koopman theorey:

\begin{theorem}[Generalization Bound for Koopman-enhanced Sequence Prediction]
\label{thm:koopman_generalization}
For a temporal prediction block of size $B$, assume the reconstruction mapping satisfies the conditional spatial stability with constant $C_{\mathrm{stab}}$ and a locally well-conditioned Encoder-Decoder mapping with constant $C_{\Psi E}$. Then, for a future inference step $n \in \{1,\dots,B-1\}$, the expected HR error is bounded by:
\[
    \mathcal{E}_{\mathrm{HR}}^{(n)}(h)
    \le
    C_{\mathrm{stab}}^2 C_{\Psi E}^2
    \|\mathbf{K}\|_2^{2n}
    \mathcal{E}_{\mathrm{LR}}^{(0)}(\mathbf{P}_{h \to l}^{\top} h)
    +
    \mathcal{O}(\epsilon_{\mathrm{proj}})
\]
where $\mathcal{E}_{\mathrm{LR}}^{(0)}(\mathbf{P}_{h \to l}^{\top} h)$ denotes the projected LR-domain error of the starting time step, and $\mathcal{O}(\epsilon_{\mathrm{proj}})$ absorbs the residual errors arising from projection mismatch, finite-dimensional Koopman approximation, and Encoder–Decoder reconstruction inaccuracies.
\end{theorem}

Applying Assumption~\ref{assum:conditional_stability} to the predicted mapping at time step $t+n$, the HR error at step $n$ satisfies:
\begin{equation}
\mathcal{E}_{\mathrm{HR}}^{(n)}(h)
\le
C_{\mathrm{stab}}^2
\mathbb{E} \left[
\|\mathbf{P}_{h \to l}^{\top}(h(\tilde{\mathbf{Q}}_l(t+n)) - h^*(\mathbf{Q}_l(t+n)))\|_F^2
\right]
\label{thm2-2}
\end{equation}
where
$
\mathcal{E}_{\mathrm{HR}}^{(n)}(h)
=
\mathbb{E}
\left[
\|h(\tilde{\mathbf{Q}}_l(t+n)) - h^*(\mathbf{Q}_l(t+n))\|_F^2
\right]
$.
To evaluate the projected spatial error at step $n$, we consider the Koopman linear transition
$
\boldsymbol{z}^K(t+n)=\mathbf{K}^n\boldsymbol{z}(t)
$
together with the ideal latent evolution
$
\boldsymbol{z}^*(t+n)=\mathbf{K}^n\boldsymbol{z}^*(t)+\boldsymbol{\eta}_n
$, where $\boldsymbol{\eta}_n$ denotes the accumulated finite-dimensional Koopman truncation error. By using the Lipschitz continuity of the decoder-projection map and the local conditioning of the encoder, we establish the upper bound of the projected spatial error at step $n$: 
\begin{equation}
\begin{aligned}
&\mathbb{E} \left[
\|\mathbf{P}_{h \to l}^{\top}(h(\tilde{\mathbf{Q}}_l(t+n)) - h^*(\mathbf{Q}_l(t+n)))\|_F^2
\right] \\
&\qquad \le
C_{\Psi E}^2
\|\mathbf{K}\|_2^{2n}
\mathbb{E} \left[
\|\mathbf{P}_{h \to l}^{\top}(h(\tilde{\mathbf{Q}}_l(t)) - h^*(\mathbf{Q}_l(t)))\|_F^2
\right]
+
\mathcal{O}(\epsilon_{\mathrm{proj}})
\end{aligned}
\label{thm2-3}
\end{equation}
where $C_{\Psi E}$ is independent of the prediction horizon $n$. The derivation of Eq.~(\ref{thm2-3}) is provided in Supplementary Material~\ref{app:n-error}.

Additionally, as established in Lemma~\ref{lem:stability_bound}, the expected initial projected structural error is upper-bounded by the expected LR loss, $\mathcal{E}_{\mathrm{LR}}^{(0)}(\mathbf{P}_{h\to l}^{\top} h)$:
\begin{equation}
\mathbb{E}\Bigl[ \|\mathbf{P}_{h\to l}^{\top}\bigl(h(\tilde{\mathbf{Q}}_l(t))-h^*(\mathbf{Q}_l(t))\bigr)\|_F^2 \Bigr] \le \mathcal{E}_{\mathrm{LR}}^{(0)}(\mathbf{P}_{h\to l}^{\top} h)
\label{thm2-5}
\end{equation}
Substituting Eq.~(\ref{thm2-5}) into Eq.~(\ref{thm2-3}) yields:
\begin{equation}
\begin{aligned}
\mathbb{E} \left[
\|\mathbf{P}_{h \to l}^{\top}(h(\tilde{\mathbf{Q}}_l(t+n)) - h^*(\mathbf{Q}_l(t+n)))\|_F^2
\right]  \le
C_{\Psi E}^2
\|\mathbf{K}\|_2^{2n}
\mathcal{E}_{\mathrm{LR}}^{(0)}(\mathbf{P}_{h\to l}^{\top} h)
+
\mathcal{O}(\epsilon_{\mathrm{proj}})
\end{aligned}
\label{thm2-6}
\end{equation}
Finally, substituting Eq.~(\ref{thm2-6}) into Eq.~(\ref{thm2-2}) yields the claimed bound in Theorem~\ref{thm:koopman_generalization}.

Theorem~\ref{thm:koopman_generalization} mathematically proves that enforcing a linearized latent space via the regularized Koopman operator serves as an effective mechanism for error mitigation. 
Specifically, the dominant temporal error propagation is governed by $\|\mathbf{K}\|_2^{2n}$. By constraining  $\|\mathbf{K}\|_F$, the framework strictly caps the expansion rate ($\|\mathbf{K}\|_2 \le \gamma < L_{\mathcal{M}}$). Thus, $\|\mathbf{K}\|_2^{2n}$ is systematically suppressed compared to the $\mathcal{O}(L_{\mathcal{M}}^{n})$ explosion characteristic of standard recurrent models. 

\section{Error Scale of Purely Data-driven Recurrent Models} \label{app:normal-error}

Let $\mathbf{Q}(t)$ denote the true state at time $t$, and $\widehat{\mathbf{Q}}(t)$ denote the model-reconstructed state at time $t$. We assume that, at the reference time $t$ from which temporal propagation starts, the prediction or spatial reconstruction error is bounded by $\epsilon(t)$:
\[
\|\widehat{\mathbf{Q}}(t) - \mathbf{Q}(t)\| \le \epsilon(t)
\]
The true system evolves according to the underlying transition operator $\mathcal{F}$, such that $\mathbf{Q}(t+1) = \mathcal{F}(\mathbf{Q}(t))$. Meanwhile, the model advances the state using the learned nonlinear function $\mathcal{M}_\theta$, yielding $\widehat{\mathbf{Q}}(t+1) = \mathcal{M}_\theta(\widehat{\mathbf{Q}}(t))$. Because $\mathcal{M}_\theta$ is an approximation of the true dynamics $\mathcal{F}$, there exists a single-step structural approximation error (or model bias) bounded by a constant $\epsilon_\text{str} \ge 0$ for any valid state:
\[
\|\mathcal{M}_\theta(\mathbf{Q}(t)) - \mathcal{F}(\mathbf{Q}(t))\| \le \epsilon_\text{str}
\]

Now, consider the discrepancy between the predicted state and the true state at the next time step, $t+1$:
\begin{equation*}
\begin{aligned}
\|\widehat{\mathbf{Q}}(t+1) - \mathbf{Q}(t+1)\| &= \|\mathcal{M}_\theta(\widehat{\mathbf{Q}}(t)) - \mathcal{F}(\mathbf{Q}(t))\|\\
&= \|\mathcal{M}_\theta(\widehat{\mathbf{Q}}(t)) - \mathcal{M}_\theta(\mathbf{Q}(t)) + \mathcal{M}_\theta(\mathbf{Q}(t)) - \mathcal{F}(\mathbf{Q}(t))\|\\
&\le \|\mathcal{M}_\theta(\widehat{\mathbf{Q}}(t)) - \mathcal{M}_\theta(\mathbf{Q}(t))\| + \|\mathcal{M}_\theta(\mathbf{Q}(t)) - \mathcal{F}(\mathbf{Q}(t))\|\\
&\le L_{\mathcal{M}} \|\widehat{\mathbf{Q}}(t) - \mathbf{Q}(t)\| + \epsilon_\text{str}
\end{aligned}
\end{equation*}
where the first inequality is true due to the triangle inequality, the last inequality is true due to that $\mathcal{M}_\theta$ is $L_{\mathcal{M}}$-Lipschitz.
Substituting the starting error bound $\epsilon(t)$ yields the error at step 1:
\[
\epsilon(t+1) \le L_{\mathcal{M}} \epsilon(t) + \epsilon_\text{str}
\]
To evaluate the error after $n$ steps, we unroll this recurrence relation,
and the accumulated error at time $t+n$ is bounded by:
\[
\epsilon(t+n) \le L_{\mathcal{M}}^n \epsilon(t) + \epsilon_\text{str} \sum_{i=0}^{n-1} L_{\mathcal{M}}^i
\]

For purely data-driven recurrent models dealing with complex, unstable nonlinear dynamics, the learned operator is typically expansive, i.e., $L_{\mathcal{M}} > 1$. As such, we have the following error bound:
\[
\epsilon(t+n) \sim \mathcal{O}(L_{\mathcal{M}}^n \epsilon(t))
\]

\section{Details of Eq.~(\ref{thm2-3})} \label{app:n-error}
To establish the inequality in Eq.~(\ref{thm2-3}), we must bridge the temporal error growth in the latent Koopman space with the spatial error in the observable LR domain. The derivation proceeds in the following three main steps: 

\paragraph{Error Propagation in the Latent Koopman Space.}
Let $\boldsymbol{z}(t) = \operatorname{vec}(\mathrm{E}_\theta(\tilde{\mathbf{Q}}_l(t)))$ denote the encoded latent state vector from the observations, and let $\boldsymbol{z}^*(t)$ be the ideal latent representation of the true system state. Based on the Koopman operator framework, the predicted latent state advances linearly over $n$ steps:
$$
\boldsymbol{z}^K(t+n) = \mathbf{K}^n \boldsymbol{z}(t)
$$
Because the finite-dimensional matrix $\mathbf{K}$ is only an approximation of the infinite-dimensional Koopman operator, this evolution introduces an accumulated truncation error $\boldsymbol{\eta}_n$:
$$
\boldsymbol{z}^*(t+n) = \mathbf{K}^n \boldsymbol{z}^*(t) + \boldsymbol{\eta}_n
$$
Subtracting the true latent trajectory from the prediction yields the latent error at step $n$:
$$
\boldsymbol{z}^K(t+n) - \boldsymbol{z}^*(t+n) = \mathbf{K}^n \bigl( \boldsymbol{z}(t) - \boldsymbol{z}^*(t) \bigr) - \boldsymbol{\eta}_n
$$
Applying the triangle inequality and the sub-multiplicativity of matrix norms yields:
\begin{equation}
\|\boldsymbol{z}^K(t+n) - \boldsymbol{z}^*(t+n)\|_2 \le \|\mathbf{K}\|_2^n \|\boldsymbol{z}(t) - \boldsymbol{z}^*(t)\|_2 + \|\boldsymbol{\eta}_n\|_2
\label{eq:latent_bound_step1}
\end{equation}
\paragraph{Connecting Latent Error to Projected Spatial Error.}
The decoder $\mathrm{D}_\theta$ further maps the latent states back to the HR space, which is then projected to the LR space via $\mathbf{P}_{h \to l}$. Let us define the composite mapping as $\Psi(\boldsymbol{z}) = \mathbf{P}_{h \to l}^{\top} \mathrm{D}_\theta(\operatorname{vec}^{-1}(\boldsymbol{z}))$. Because neural networks with standard activation functions (such as ELU) and bounded weights are Lipschitz continuous, the composite function $\Psi$ possesses a Lipschitz constant $L_\Psi > 0$. Therefore, the spatial error at step $n$ is bounded by the latent error at step $n$:
\begin{equation}
\|\mathbf{P}_{h \to l}^{\top}(h(\tilde{\mathbf{Q}}_l(t+n)) - h^*(\mathbf{Q}_l(t+n)))\|_F \le L_\Psi \|\boldsymbol{z}^K(t+n) - \boldsymbol{z}^*(t+n)\|_2 + \epsilon_{D}
\label{eq:lipschitz_decoder}
\end{equation}
where $\epsilon_{D}$ represents the intrinsic structural approximation error of the decoder.

Similarly, assuming the learned latent space is a well-conditioned observable space (bi-Lipschitz mapping), the initial latent error at $t$ is bounded by the corresponding projected spatial error scaled by a constant $C_E$:
\begin{equation}
\|\boldsymbol{z}^K(t) - \boldsymbol{z}^*(t)\|_2 \le C_E \|\mathbf{P}_{h \to l}^{\top}(h(\tilde{\mathbf{Q}}_l(t)) - h^*(\mathbf{Q}_l(t)))\|_F + \epsilon_{E}
\label{eq:lipschitz_encoder}
\end{equation}

\paragraph{Synthesis and Expectation.}
Substituting Eq.~(\ref{eq:lipschitz_encoder}) into Eq.~(\ref{eq:latent_bound_step1}), and then substituting the resulting bound into Eq.~(\ref{eq:lipschitz_decoder}), we relate the future projected spatial error directly to the current projected spatial error:
\[
\|\mathbf{P}_{h \to l}^{\top}(h(\tilde{\mathbf{Q}}_l(t+n)) - h^*(\mathbf{Q}_l(t+n)))\|_F
\le
C_{\Psi E}\|\mathbf{K}\|_2^n
\|\mathbf{P}_{h \to l}^{\top}(h(\tilde{\mathbf{Q}}_l(t)) - h^*(\mathbf{Q}_l(t)))\|_F
+
\epsilon_{\text{total}}
\]
where $C_{\Psi E}=L_\Psi C_E$ is a finite constant determined by the decoder Lipschitz constant and the encoder conditioning, and $\epsilon_{\text{total}}$ aggregates the scaled Koopman truncation error $\boldsymbol{\eta}_n$, encoder error $\epsilon_E$, and decoder error $\epsilon_D$. Squaring both sides and taking the expectation over the data distribution gives:
\[
\mathbb{E} \left[
\|\mathbf{P}_{h \to l}^{\top}(h(\tilde{\mathbf{Q}}_l(t+n)) - h^*(\mathbf{Q}_l(t+n)))\|_F^2
\right]
\le
C_{\Psi E}^2 \|\mathbf{K}\|_2^{2n}
\mathbb{E} \left[
\|\mathbf{P}_{h \to l}^{\top}(h(\tilde{\mathbf{Q}}_l(t)) - h^*(\mathbf{Q}_l(t)))\|_F^2
\right]
+
\mathcal{O}(\epsilon_{\mathrm{proj}})
\label{thm2-3_derived}
\]
Here, $\mathcal{O}(\epsilon_{\mathrm{proj}})$ absorbs the expected contribution of the residual structural and approximation errors. The constant $C_{\Psi E}^2$ does not depend on the prediction horizon $n$; therefore, the temporal growth rate of the error is governed primarily by $\|\mathbf{K}\|_2^{2n}$. This completes the derivation of the error scaling in Eq.~(\ref{thm2-3}).

\section{Summary of Notations} \label{app:notations}
 Throughout this paper, we adopt strict typographical conventions to explicitly distinguish mathematical entities by their dimensionality:
\begin{itemize}
    \item \textbf{Matrices and Tensors} (2D arrays and higher) are denoted by bold, upright uppercase letters (e.g., $\mathbf{Q}, \mathbf{K}, \mathbf{W}$).
    \item \textbf{Vectors} (1D arrays) are denoted by bold, italic lowercase letters (e.g., $\boldsymbol{q}, \boldsymbol{z}, \boldsymbol{w}$).
    \item \textbf{Scalars and Individual Elements} (0D values) are denoted by standard italic letters (e.g., $u, v, \epsilon$).
\end{itemize}

\renewcommand{\arraystretch}{0.75}

\begin{longtable}{@{} p{0.2\linewidth} >{\raggedright\arraybackslash}p{0.85\linewidth} @{}}
  \label{tab:notations} \\
  \toprule
  \textbf{Symbol} & \textbf{Description} \\
  \midrule
  \endfirsthead
  
  \multicolumn{2}{c}%
  {{\bfseries \tablename\ \thetable{} -- continued from previous page}} \\
  \toprule
  \textbf{Symbol} & \textbf{Description} \\
  \midrule
  \endhead
  
  \midrule 
  \multicolumn{2}{r}{{Continued on next page}} \\
  \endfoot

  \bottomrule
  \endlastfoot

$\mathcal{T}, N_\text{t}$ & Sequence of time points and number of time steps. \\
$\mathcal{X}_l, N_\text{s}$ & Set of LR spatial locations and its cardinality. \\
$\mathcal{X}_h, N_\text{s}^*$ & Set of HR spatial locations and its cardinality. \\
$\mathcal{B}_l, B$ & Short temporal block of LR observations and its size. \\
$N$ & Number of vertices in a generic graph $\mathcal{G}$. \\

$u, v$ & Normalized transmembrane potential and recovery variable. \\
$C_1, C_2, \alpha, e_0, \mu_1, \mu_2$ & Parameters of the Aliev--Panfilov (AP) cardiac model. \\
$e_1, e_2$ & Diffusion coefficients for $u$ and $v$. \\
$g_1(u,v), g_2(u,v)$ & Nonlinear reaction terms of the AP model. \\
$M$ & Complex 3D surface geometry manifold. \\
$\Delta_M, \nabla_M$ & Laplace--Beltrami operator (surface Laplacian) and surface gradient on $M$. \\
$\mathbf{n}$ & Outward unit normal vector at the surface boundary. \\

$\boldsymbol{\xi}(\boldsymbol{x}, t), \sigma_\xi$ & Gaussian measurement noise vector and its standard deviation. \\
$\mathbf{\Xi}_\text{mea}(t)$ & Matrix of zero‑mean measurement noise corrupting LR observations. \\
$\sigma_{\text{mea}}^2$ & Variance of the measurement noise $\mathbf{\Xi}_\text{mea}$. \\

$\mathcal{G}, \mathcal{V}, \mathcal{E}$ & Undirected graph, its vertex set, and adjacency matrix. \\
$\mathbf{W}, \boldsymbol{w}(i,j)$ & Edge attribute tensor and the normalized spatial displacement vector between vertices $i$ and $j$. \\
$\mathcal{N}(i)$ & Local geometric neighborhood of node $i$. \\
$\mathbf{P}_k$ & Cluster assignment matrix for hierarchical graph pooling/unpooling at level $k$. \\
$\boldsymbol{\Delta}_{k+1}$ & Spatial normalization diagonal matrix used during hierarchical coarsening. \\
$\mathbf{P}_{h \to l}$ & Linear projection matrix mapping the HR mesh to the LR observation nodes. \\

$\mathrm{E}_\theta, \mathrm{D}_\theta$ & Spatial encoder and decoder networks with trainable parameters $\theta$. \\
$C$ & Number of feature channels in the system state. \\
$C_\text{in}, C_\text{out}$ & Input and output feature dimensions for a graph convolution layer. \\
$\mathbf{G}_{\mathbf{\Theta}}(\boldsymbol{w})$ & Continuous graph convolution kernel matrix (size $C_\text{out}\times C_\text{in}$) parameterized by edge attributes $\boldsymbol{w}$. \\
$\mathbf{\Theta} = \{\mathbf{\Theta}_{\boldsymbol{p}}\}$ & Set of trainable parameter matrices for the B‑spline convolution kernel. \\
$B_{\boldsymbol{p}}(\boldsymbol{w})$ & Tensor‑product B‑spline basis function evaluated at edge attribute $\boldsymbol{w}$ for basis index $\boldsymbol{p}$. \\
$N_{i,p_i}(w_i)$ & 1D B‑spline basis function of specified degree, evaluated at coordinate $w_i$. \\
$\boldsymbol{r} = [r_1,\dots,r_D]^\top$ & Kernel resolution vector specifying the number of basis functions per pseudo‑coordinate dimension. \\
$\mathcal{P}$ & Cartesian product of the B‑spline basis indices across all $D$ dimensions. \\
$\rho(\cdot)$ & ELU (Exponential Linear Unit) activation function. \\

$\boldsymbol{q}(\boldsymbol{x}, t)$ & Reference (ground‑truth) vectorized features at location $\boldsymbol{x}$ and time $t$. \\
$\hat{\boldsymbol{q}}(\boldsymbol{x}, t)$ & Predicted vectorized features. \\
$\tilde{\boldsymbol{q}}(\boldsymbol{x}, t)$ & Noisy physical measurements used as input. \\
$\boldsymbol{q}^{(k)}$ & Vectorized node features at layer $k$ within the residual S‑F Box. \\
$\mathbf{Q}_k$ & Feature matrix at the $k$‑th hierarchical graph coarsening level. \\
$\mathbf{Q}_k^\text{pool}, \mathbf{Q}_k^\text{unpool}$ & Feature matrices after pooling/unpooling operations at level $k$. \\
$\mathcal{Q}_l, \tilde{\mathcal{Q}}_l, \mathcal{Q}_h$ &  Full spatiotemporal tensors for noise-free LR, noisy LR and reconstructed HR states. \\
$\mathbf{Q}_l(t), \tilde{\mathbf{Q}}_l(t)$ &  Feature matrix of the entire noise-free or noisy LR space at time $t$. \\
$\widehat{\mathbf{Q}}_h(t)$ & Reconstructed HR feature matrix at time $t$ (direct autoencoder output). \\
$\widehat{\mathbf{Q}}_h^K(t)$ & Predicted HR feature matrix obtained by advancing the latent state via the Koopman operator. \\

$\mathcal{G}_z, N_{s,z}, C_z$ & Latent graph after encoding, its number of nodes, and its feature dimension. \\
$\mathbf{Z}(t), \boldsymbol{z}(t)$ & Latent feature matrix and its flattened observable state vector ($\boldsymbol{z} = \operatorname{vec}(\mathbf{Z})$). \\
$\boldsymbol{z}^*(t), \boldsymbol{z}^K(t)$ & True latent state vector and predicted latent state vector advanced via $\mathbf{K}$. \\
$\mathbf{K}$ & Trainable matrix serving as a finite‑dimensional approximation of the Koopman operator. \\
$d$ & Dimension of the flattened observable state vector ($d = N_{s,z} C_z$). \\
$D$ & Dimensionality of the pseudo‑coordinate space (e.g., $D=3$ for 3D geometries). \\

$\mathcal{L}, \mathcal{L}_{\mathrm{d}}, \mathcal{L}_{\mathrm{phy}}$ & Total loss, data‑driven reconstruction loss, and physics‑based residual loss. \\
$w_{\mathrm{phy}}$ & Physics penalty weight modulating the contribution of $\mathcal{L}_{\mathrm{phy}}$. \\
$\mathcal{R}_u, \mathcal{R}_v$ & Physics residuals quantifying PDE violations for variables $u$ and $v$. \\
$RE_u, RE_v, RE_{\text{total}}$ & Relative errors for $u$, $v$, and their aggregated metric. \\
$\eta$ & Learning rate for gradient‑based optimization. \\

$h^*$ & Unknown true HR mapping that exactly satisfies the governing PDEs. \\
$\mathcal{R}(h^*)$ & Expected physics residual of the true HR mapping (identically zero). \\
$h$ & Parameterized candidate mapping for HR reconstruction. \\
$\mathcal{E}_{\text{HR}}(h)$ & Expected HR reconstruction error (squared Frobenius norm). \\
$\|\cdot\|_F$ & Frobenius norm of a matrix or tensor. \\
$\langle \cdot, \cdot \rangle_F$ & Frobenius inner product. \\
$\mathcal{H}$ & Hypothesis space of all functions representable by the unconstrained P‑K‑GCN. \\
$\mathcal{H}_{\mathrm{phy}}(\epsilon)$ & Physics‑constrained hypothesis space with PDE residual tolerance $\epsilon$. \\
$\epsilon$ & Tolerance parameter defining $\mathcal{H}_{\mathrm{phy}}(\epsilon)$. \\
$\mathcal{H}_l$ & Induced hypothesis class in the LR observational space. \\
$h_l$ & Mapping in the induced LR class ($h_l = \mathbf{P}_{h\to l}^\top h$). \\
$\mathcal{H}_{l,\mathrm{phy}}(\epsilon)$ & Physics‑constrained induced hypothesis class in the LR space. \\
$\hat{\mathfrak{R}}_{\tilde{\mathcal{Q}}_l}(\mathcal{H}_l)$ & Empirical Rademacher complexity of the LR hypothesis class on noisy observations $\tilde{\mathcal{Q}}_l$.  \\
$\sigma_i, \boldsymbol{\sigma}$ & Rademacher random variables taking values $\pm1$ with equal probability. \\
$c$ & Clipping bound for the data‑driven loss in Rademacher analysis. \\
$\delta$ & Confidence parameter for uniform convergence bounds. \\
$\mathcal{E}_{\text{LR}}(h_l)$ & Expected low‑resolution risk (population loss over LR observations). \\
$\hat{\mathcal{E}}_{\text{LR}}(h_l)$ & Empirical low‑resolution risk computed on the training block. \\
$C_{\text{stab}}$ & Conditional stability constant linking HR deviations to their LR projections. \\

$\mathcal{F}$ & True (unknown) nonlinear transition operator governing temporal evolution. \\
$\mathcal{M}_\theta$ & Parameterized nonlinear transition function learned by standard recurrent models. \\
$L_{\mathcal{M}}$ & Local Lipschitz constant of $\mathcal{M}_\theta$. \\
$\epsilon(t)$ & Initial spatial reconstruction error or measurement noise at time $t$. \\
$\mathcal{H}_{\text{temp}}$ & Unconstrained temporal hypothesis space of arbitrary nonlinear recurrent mappings. \\
$\mathcal{H}_{\text{Koop}}$ & Koopman‑constrained temporal space restricted to linear latent transitions via $\mathbf{K}$. \\
$e_z(t)$ & Initial latent representation error $\|\boldsymbol{z}(t) - \boldsymbol{z}^*(t)\|_2$. \\
$\boldsymbol{\eta}_n$ & Accumulated approximation error of the true latent state evolution after $n$ steps. \\
$\epsilon_{\mathrm{proj}}$ & Cumulative projection and linearization error in the Koopman temporal bound. \\
$\mathcal{E}_{\mathrm{HR}}^{(n)}(h)$ & Expected HR spatiotemporal error for a future inference step $n$. \\
$\mathcal{E}_{\mathrm{LR}}^{(0)}$ & Empirical LR error at the initial time step $t$. \\

$\mathcal{L}^\text{Hil}$ & Hilbert space of observables (e.g., $L^2$ over the dynamical manifold). \\
$\mathcal{K}$ & Infinite‑dimensional Koopman operator acting on observables. \\
$\mathbf{F}$ & True evolution operator in the original state space. \\
$\mathcal{M}_\text{o}$ & Dynamical state space manifold. \\
$\boldsymbol{\psi}$ & Vector‑valued observable function (the encoder mapping). \\
$\mathcal{P}_{\mathcal{L}^\text{Hil}_d}$ & Orthogonal projection onto the finite‑dimensional observable subspace. \\
$\mathbf{K}_{\text{proj}}$ & Galerkin projection of $\mathcal{K}$ onto the observable subspace. \\
$\varepsilon_{\text{proj}}$ & Projection error $\|\mathcal{K}g - \mathcal{P}_{\mathcal{L}^\text{Hil}_d}\mathcal{K}g\|$. \\
$L_{\Psi}$ & Lipschitz constant of the composite spatial mapping $\Psi$. \\
$C_E$ & Lipschitz constant relating initial latent error to projected spatial error. \\
$\epsilon_D, \epsilon_E$ & Intrinsic structural approximation errors of the decoder and encoder, respectively. \\
$\epsilon_{\text{total}}$ & Aggregated residual error term in the temporal stability derivation. \\
\end{longtable}

\end{document}